\documentclass{article}

\PassOptionsToPackage{numbers, compress}{natbib}

 \usepackage{amsmath} 
  \usepackage{graphicx}

 \usepackage[preprint]{neurips_2026}

\usepackage{iftex}
\ifPDFTeX
  \usepackage[utf8]{inputenc}
\fi
\usepackage[T1]{fontenc}    %
\usepackage{xurl}           %
\usepackage{booktabs}       %
\usepackage{amsfonts}       %
\usepackage{nicefrac}       %
\usepackage{microtype}      %
\usepackage{xcolor}         %
\usepackage{float}
\usepackage{placeins}
\usepackage{hyperref}
\title{Data Density Scaling Laws for Image Self-Distillation}

\author{Alvard Barseghyan \quad Ani Vanyan \quad Hakob Tamazyan \quad Hrant Khachatrian\\
YerevaNN and Yerevan State University, Yerevan, Armenia\\
\texttt{\{alla,ani,hakob,hrant\}@yerevann.com}}

\begin{document}

\maketitle

\begin{abstract}
    Current scaling law formulations suggest that increasing the number of unique samples in a dataset consistently improves model performance, while repeated exposure to the same samples provides diminishing value. In this work, we challenge this assumption for image self-distillation models.
    We train iBOT models on random subsets of seven datasets spanning satellite imagery, driving scenes, egocentric video, and natural images, and evaluate their frozen representations on downstream tasks. We propose a density-interaction scaling law that combines unique-image count $U$ and repetition $R$, fitting it on runs with at most 50 million image presentations and testing it at larger budgets $C$. On Maxar dataset, a random subset containing approximately 1.5\% of the training pool improves average nearest-neighbor classification accuracy by 3.11 percentage points over training on the full pool at the same budget of 200 million image presentations. Smaller subsets also outperform the full pool on SATLAS and Ego4D, whereas ImageNet favors larger pools at higher budgets. At a fixed subset size, further training can improve performance, lead to a plateau, or reduce performance. Our law captures smaller-subset advantages that the compared baselines cannot represent and improves prediction at higher budgets on Maxar, SATLAS, and Ego4D under nearest-neighbor evaluation. These findings show that both the number of unique images and how often they are repeated matter when allocating a training budget.

\end{abstract}
\section{Introduction}
\label{intro}

Recent progress in foundation models has been strongly associated with scaling both model size and dataset size \cite{Radford2019LanguageMA, brown2020language, chinchillalaw}. In language modeling, empirical scaling laws describe how performance changes with model parameters, training data, and compute \cite{kaplan2020scaling, chinchillalaw}. Similar trends have motivated large-scale representation learning in computer vision \cite{mae, beit3, dinov2, dinov3}. However, with a fixed training budget, is it better to train on more images or repeat a smaller set more often?

This question matters for datasets such as satellite imagery and driving video, where many images show similar scenes. Adding more images may therefore add relatively little new information. Data-constrained scaling laws distinguish unique samples from repeated exposures and assign less value to each additional repetition \cite{huggingfacelaw}. In image self-distillation, repeated images can produce different augmented views, while the teacher providing the training targets also changes during training \cite{ibot, dinov2}. These differences motivate studying whether existing scaling laws describe the effects of repetition in this setting.

We study this question across seven pretraining datasets: Maxar \cite{maxar}, SATLAS \cite{satlas}, Waymo \cite{waymo_e2e}, nuPlan \cite{nuplan}, Ego4D \cite{ego4d}, ImageNet \cite{imagenet}, and YFCC15M \cite{yfcc100m,clip,openai_yfcc_subset}. Our main experiments use iBOT with the same backbone and training settings across subset sizes within each dataset. We vary the \textit{number of selected images} $U$ and the \textit{training budget} $C$, measured as the total number of source-image presentations during the training. We use $R$ to denote how many times each image is seen during training, on average. We call this the \textit{repetition density}. We measure downstream performance using frozen representations, with kNN and logistic regression evaluated separately. Our main analysis compares random subsets with the full training pool.

Our experiments reveal several patterns. At the same training budget, using fewer unique images can improve performance, have little effect, or reduce performance, depending on the dataset. On Maxar, a random subset containing only 1.5\% of the training pool improves average downstream kNN performance by 3.11 percentage points compared with the full training pool, using the same budget of 200M image presentations. Smaller-subset advantages also occur on SATLAS and Ego4D, while ImageNet benefits from larger image pools at higher budgets and Waymo is comparatively insensitive to pool size. The best evaluated subset can also change with the training budget. Finally, when the image pool is fixed, further repetition can improve performance, lead to a plateau, or reduce performance. Together, these observations show why both dataset size and training duration matter when choosing how much data to use. 

To study whether scaling laws can predict these patterns, we adapt three formulations from language modeling: compute-only, data-constrained, and prescriptive \cite{chinchillalaw,huggingfacelaw,lovelace2026prescriptive}, to predict downstream performance in image self-distillation at a fixed model size. The compute-only law does not distinguish between subset sizes at the same training budget. The data-constrained law models diminishing benefits from repetition, while the prescriptive law also allows repetition to reduce performance. However, none can predict that a smaller subset performs better than a larger one at the same budget. We therefore extend the compute-only law with a term that combines the number of unique images seen during training $U_{seen}$ and their repetition $R$, giving our density-interaction law. We fit all four laws using low-budget experiments and evaluate their predictions at higher budgets. The benefits of the proposed law vary across datasets and evaluation protocols.

Our main contributions are:
\begin{itemize}
    \item We provide a systematic study of dataset size and repetition in image self-distillation across seven datasets, showing when smaller random subsets outperform the full training pool and how these results change with training budget.
    \item We propose a density-interaction scaling law and evaluate its ability to predict downstream performance using low-budget training results.
    \item We show that the subset advantage seen on Maxar consistently holds across model sizes, self-distillation methods, and subset selection strategies. %
\end{itemize}

\section{Related Work}
\label{related}

\textbf{Scaling laws and repeated data.}
Scaling laws describe how model performance changes with model size, training data, and compute. Hoffmann et al.~\cite{chinchillalaw} study how to choose model size and the number of training tokens for a fixed compute budget. Muennighoff et al.~\cite{huggingfacelaw} extend this approach to settings with limited data. Their law distinguishes unique tokens from repeated exposures and assigns less value to each additional repetition. Their experiments show that repeating data for several passes can perform similarly to using fresh data, but the benefit decreases with further repetition. We study whether this law also describes image self-distillation. In our comparison, its effective-data formulation cannot predict that a smaller subset outperforms a larger one at the same model size and compute budget.

In parallel work, Lovelace et al.~\cite{lovelace2026prescriptive} add an overfitting penalty to model the harmful effects of repetition in language-model training. Their findings overlap with ours: repetition can hurt performance, rather than only give smaller gains. We adapt their law as a baseline and test whether it captures the behavior observed in our experiments.

\textbf{Self-distillation.}
Visual self-distillation trains a student network to match a teacher's predictions for different views of an image. iBOT~\cite{ibot} combines this approach with masked patch prediction. DINOv2~\cite{dinov2} uses related objectives to learn visual features from large image collections. We use iBOT for our main experiments and DINOv2 to test whether the smaller-subset advantage also appears with another method.

\textbf{Dataset size in self-supervised learning.}
Previous work shows that more pretraining data does not help equally in every setting. El-Nouby et al.~\cite{elnouby2021largescale} show that denoising autoencoders can learn useful representations from small target-task datasets. Xie et al.~\cite{xie2023datascaling} find that the benefit of more data in masked image modeling depends on model size and training duration. Lu et al.~\cite{lu2023delving} study random subsets of web images and find that the choice of pretraining data often has a limited effect on masked image modeling. We study image self-distillation across seven datasets. We vary dataset size and training budget, then evaluate the learned features using kNN and logistic regression separately.

\textbf{Data selection and redundancy.}
Data selection methods aim to reduce training costs or improve performance by choosing which examples to keep. Sorscher et al.~\cite{sorscher2022beyond} study how data pruning changes scaling behavior. SemDeDup~\cite{semdedup} removes semantically similar examples using pretrained embeddings. Goyal et al.~\cite{scalinglawsfordatafiltering} show that data filtering should depend on the training budget. A filtered subset can perform well at low compute but lose its advantage as its examples are repeated.

We compare methods that select images based on balance, diversity, or quality. Hierarchical sampling~\cite{hierarchical} balances the selected images across clusters. DASSOT~\cite{van2024graph} selects a balanced subset using similarities between examples. NIQE~\cite{niqe} estimates image quality from natural-image statistics. We compare these methods with random sampling at the same subset sizes and training budgets. Our main scaling analysis uses random subsets and the original set, so it does not depend on a particular selection method.

\textbf{Sample interactions.}
A sample's effect can depend on the other samples used for training. Hu et al.~\cite{influence} show that the effect of a group of samples cannot always be predicted by adding their individual effects. For contrastive self-supervised learning, Joshi and Mirzasoleiman~\cite{joshi2023dataefficient} show that a sample's usefulness depends on how its augmented views relate to those of other samples. These findings motivate our hypothesis that some combinations of samples may make learning less effective. They do not directly establish this effect in self-distillation.

\section{Experimental Design}
\subsection{Pretraining Datasets}
We pretrain independently on seven datasets covering four visual domains. For each dataset, the full training pool contains $F$ images and is denoted by Full. This pool may differ from the complete source dataset. A selected subset contains $U\leq F$ images; training on Full uses $U=F$.

\textbf{Remote sensing:} The Maxar collection \citep{maxar} contains $F=51{,}197{,}237$ satellite-image tiles, while SATLAS \citep{satlas} contains $F=10{,}969{,}092$ images. We use satellite imagery as our primary experimental domain because repeated observations of the same geographic regions under different seasonal, atmospheric, and illumination conditions produce substantial near-duplicate and semantically overlapping content. This redundancy motivates testing whether repeated training on smaller image pools can improve representation quality under a fixed training budget.

\textbf{Driving:} Our Waymo~\citep{waymo_e2e} and nuPlan~\citep{nuplan} collections contain $F=5{,}806{,}392$ and $F=25{,}481{,}542$ images, respectively. Like satellite imagery, driving datasets contain repeated observations of related scenes: temporally adjacent frames, repeated routes, recurring road layouts, and similar weather, lighting, and traffic conditions. They provide a complementary domain for testing whether the subset--repetition trade-offs observed in remote sensing extend to ground-level visual data.

\textbf{Egocentric video:} The Ego4D collection~\citep{ego4d} contains $F=38{,}387{,}515$ frames depicting everyday activities from a first-person perspective. It extends our study to another video domain, allowing us to examine the subset--repetition trade-off beyond driving scenes.

\textbf{Natural images:} ImageNet-1K \citep{imagenet,ilsvrc} contains $F=1{,}281{,}167$ training images in a curated, object-centered collection organized around semantic categories. It provides a contrasting setting to satellite imagery and the driving datasets: reducing the selected pool can remove useful coverage of objects and their appearances. We use it to test whether smaller-subset advantages also occur when unique-image coverage is particularly relevant to recognition. Our YFCC15M pool contains $F=14{,}825{,}236$ images derived from the OpenAI-curated YFCC100M subset released with CLIP~\citep{yfcc100m,clip,openai_yfcc_subset} and provides a second natural-image corpus for examining how the observed trade-off varies across datasets.

\begin{figure}[h!]
    
    \centering
    \includegraphics[width=0.9\textwidth]{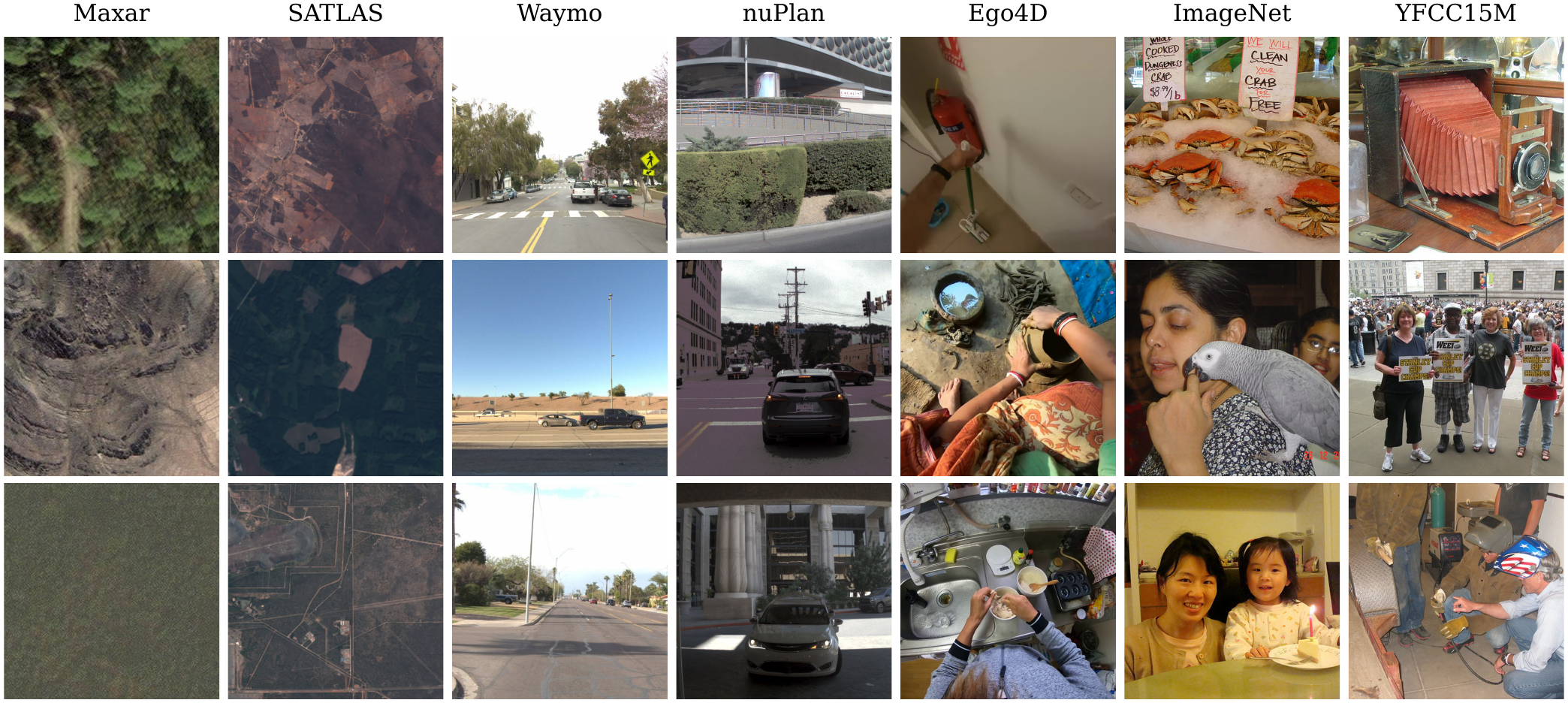}
    \caption{Samples from pretraining datasets.}
    \label{fig:pretraing_datasets}
\end{figure}

\subsection{Sampling Strategies}
\label{sec:sampling_strategies}
The experimental design serves two purposes. First, we vary the unique-data support $U$ using uniformly sampled random subsets to study its interaction with repetition at fixed training budgets. We define encountered support as $U_{\mathrm{seen}}=\min(C,U)$ and repetition as $R=C/U_{\mathrm{seen}}$. Second, we compare different subset-selection methods at matched subset sizes and training budgets to examine whether the observed subset advantage depends on the selection strategy.

\textbf{Uniform random sampling.} For each of the seven pretraining datasets, we sample images uniformly without replacement from its original experimental pool. For each dataset, we primarily consider subsets containing 100\%, 75\%, 50\%, 25\%, 12.5\%, and 6.25\% of the original pool. The main scaling-law analysis uses random subsets and Full, with dataset-specific pool sizes and available training budgets shown in Figure~\ref{fig:pretraining_setup}.

\textbf{Hierarchical sampling.} We construct subsets using the hierarchical clustering and balanced sampling approach of Vo et al.~\cite{hierarchical}. We extract [CLS] embeddings with a pretrained DINOv2 model \cite{dinov2}, then apply hierarchical $k$-means clustering across four levels with $k\in\{50\text{k},10\text{k},5\text{k},1\text{k}\}$. Balanced sampling across clusters reduces the dominance of densely populated clusters.

\textbf{DASSOT selection.} DASSOT (Data Sub-Sampling with Optimal Transport) formulates diverse subset selection as a graph-matching problem in a feature space~\cite{van2024graph}. We adapt this method through a two-stage procedure. First, we partition the original experimental pool using $k$-means clustering of pretrained DINOv2 [CLS] embeddings. We then apply a greedy DASSOT approximation independently within each cluster and combine the selected images to form the final subset. We consider this a deduplication algorithm based on cosine similarity. The cluster allocation and selection procedure are described in Appendix~\ref{app:dassot_selection}.

\textbf{NIQE-based selection.} We use the Natural Image Quality Evaluator (NIQE)~\cite{niqe}, a raw image quality metric based on natural scene statistics, to select images according to their estimated perceptual quality. Lower NIQE scores indicate better quality under the metric. This strategy provides a quality-based comparison to uniform sampling, deduplication, and representation-based selection.

We compare all four selection methods on 0.75M-image Maxar subsets at matched training budgets, using Full as a reference.

\subsection{Model Pretraining}
\begin{figure}[h!]
\centering
\includegraphics[width=\textwidth]{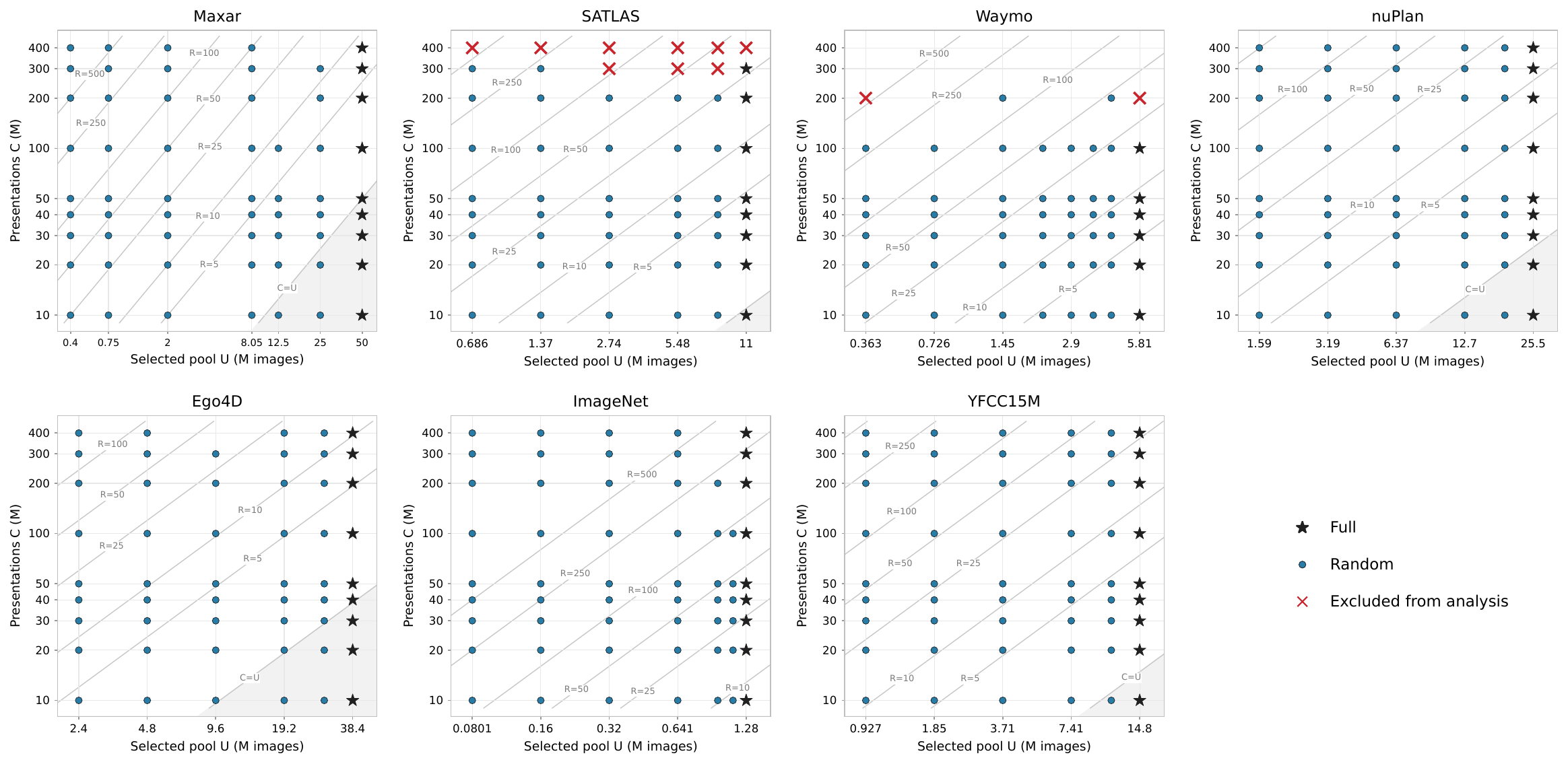}
\caption{
Pretraining grids for Maxar, SATLAS, Waymo, nuPlan, Ego4D, ImageNet, and YFCC15M. Each marker represents a selected pool size $U$ and training budget $C$, measured in millions of unique images and source-image presentations, respectively. Blue circles denote random subsets; black stars denote Full training pools. Red crosses indicate unstable runs. Repetition is defined as $R=C/\min(C,U)$. Shaded regions indicate partial passes ($C<U$), where $R=1$; the diagonal $C=U$ marks completion of the first pass. Other diagonal guides indicate constant repetition.
}
\label{fig:pretraining_setup}
\end{figure}

We pretrain iBOT with a ViT-B/16 backbone following the standard protocol \cite{ibot}. Within each pretraining dataset, we retain the same training configuration across subset sizes. Optimization is performed with AdamW using a per-GPU batch size of $200$ across $8$ H100 GPUs, resulting in a global batch size of $B=1600$. All experiments are run in distributed mode with FlashAttention-2 \citep{dao2023flashattention} for efficiency. We use a warmup--stable--decay learning-rate schedule~\citep{wsd}, with linear decay over the final 10\% of each target budget. Runs with different training budgets share the same initial training, then branch off for their final learning-rate decay. Training schedules and hyperparameter-search details are reported in Appendices~\ref{app:training_schedule} and~\ref{app:optuna_search}.
Figure~\ref{fig:pretraining_setup} illustrates the available random-subset and Full configurations across the seven datasets and compute budgets.

Given batch size $B=1600$, the number of optimization steps is $S = C/B$ (e.g., $10$M exposures $\approx 6{,}250$ steps, $50$M $\approx 31{,}250$ steps).

\subsection{Downstream Evaluation}
\label{sec:downstream_evaluation}
For classification, we evaluate $\ell_2$-normalized final-layer [CLS] embeddings from the frozen teacher backbone using cosine-similarity kNN with unweighted majority voting and logistic regression. For each pretraining group and classification task, we select $k\in\{1,5,10,30,50\}$ by mean accuracy of Full-50M. This choice remains fixed across models and budgets. Logistic regression fits an $\ell_2$-regularized classifier on the downstream training split using Cyanure~\citep{mairal2019cyanure}. Regularization is selected using Full-50M reference scores. Details appear in Appendix~\ref{app:downstream_protocol}.

\textbf{Remote sensing:} Maxar and SATLAS share six classification tasks: Brick Kiln, EuroSAT, RESISC45, UCMerced, PatternNet, and AID.

\textbf{Driving:} Waymo and nuPlan share BDD100K scene and weather classification and ACDC condition classification. Their kNN aggregates additionally include BDD100K MOT top-1 instance- and category-association accuracies, computed by matching object features across video frames. These tracking scores are excluded from logistic-regression results.

\textbf{General and egocentric recognition:} Ego4D, ImageNet, and YFCC15M share 24 classification tasks, and Ego4D additionally uses two EPIC-KITCHENS-100 noun-recognition tasks, based on single-frame and five-frame pooled features.

We average all downstream evaluation scores for each configuration, keeping kNN and logistic regression separate. The main results use kNN; logistic-regression results are reported in Appendix~\ref{app:logres_scaling}. Task details and fixed $k$ values appear in Appendix~\ref{app:downstream_tasks}.

\section{Empirical Observations}
\label{empirical_observations}
\begin{figure*}
    \centering
    \includegraphics[
        width=\textwidth,
    ]{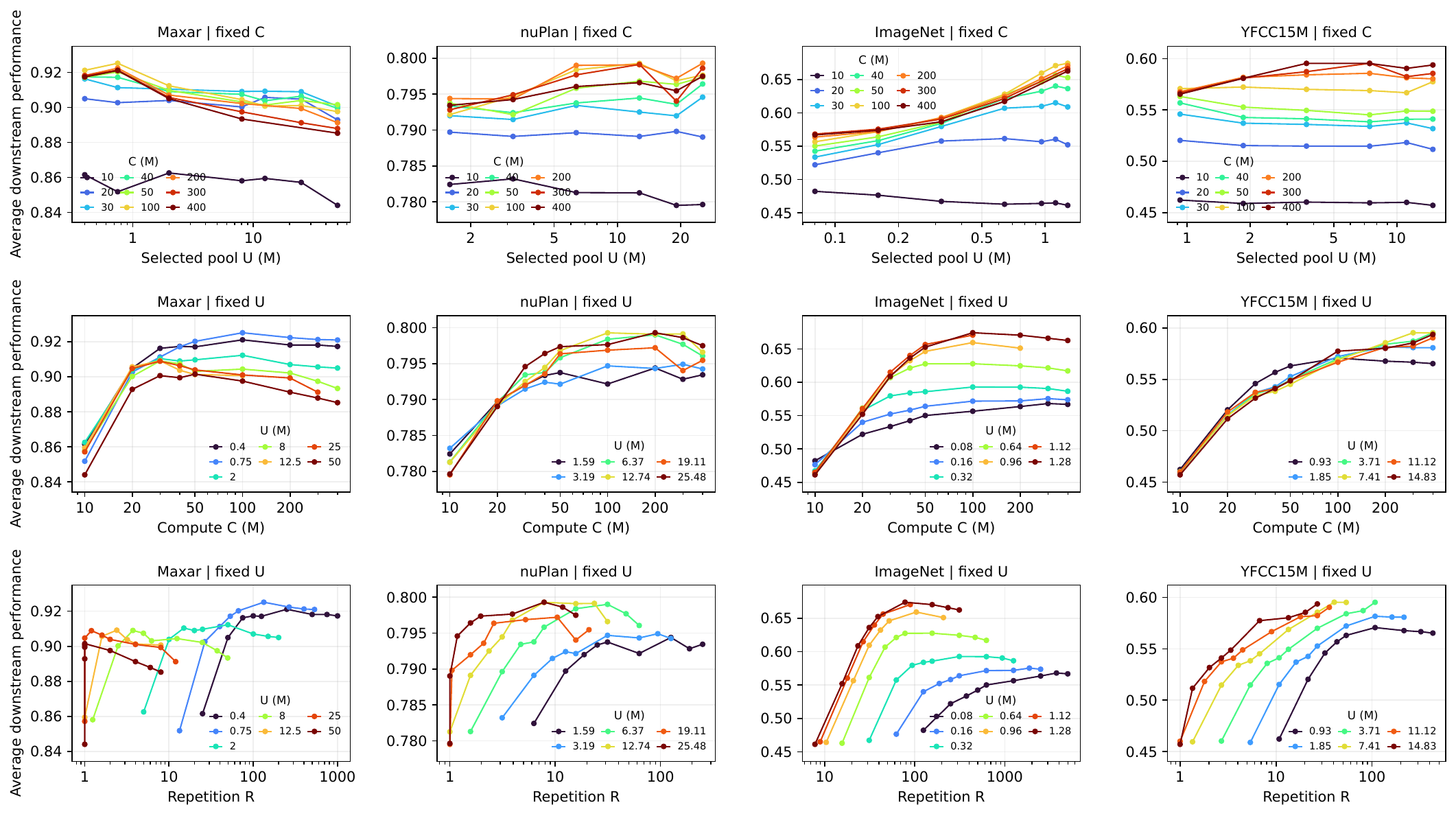}
    \caption{
    Unique-support and repetition trade-offs under kNN evaluation. Columns correspond to Maxar, nuPlan, ImageNet, and YFCC15M. The top row shows average downstream performance versus selected pool size $U$ at fixed compute $C$; the middle row shows performance versus $C$ at fixed $U$; and the bottom row shows performance versus repetition $R=C/U_{\mathrm{seen}}$, where $U_{\mathrm{seen}}=\min(C,U)$, at fixed $U$. Points represent Full ($U=F$) and random-subset observations; lines connect observations. Colors indicate compute in the top row and selected pool size in the other rows. Horizontal axes are logarithmic, and image counts are in millions.}
    \label{fig:empirical_observations_knn}
\end{figure*}

Figure~\ref{fig:empirical_observations_knn} shows how downstream kNN performance changes with dataset size and training budget. We compare random subsets and Full across seven pretraining datasets. The runs excluded from these comparisons are listed in Appendix~\ref{app:run_exclusions}. 
The corresponding logistic-regression curves appear in Appendix~\ref{app:logres_density}. At each budget, we call the highest-scoring evaluated subset, including Full, the \emph{best evaluated subset}.

\textbf{Observation 1: smaller random subsets can outperform Full.}
Maxar is the clearest case. At matched compute, Full training underperforms much smaller random subsets that are repeated many times. At $C=200$M, the random subset with $U=0.015F$ reaches 92.24\% average kNN performance, compared with 89.12\% for the Full pool. At $C=400$M, the corresponding values are 92.10\% and 88.53\%. Thus, a subset containing only 1.5\% of Full improves performance by 3.11 and 3.57 percentage points at these budgets. SATLAS and Ego4D also exhibit a smaller-subset advantage, although the gains are smaller. For example, at $C=300$M, the smallest evaluated SATLAS subset, containing 6.25\% of Full, exceeds Full by 1.13 percentage points. At $C=400$M, the best evaluated Ego4D subset, also containing 6.25\% of Full, exceeds Full by 1.10 percentage points. Under logistic regression at 200M presentations, the random 0.75M-image Maxar subset outperforms Full by 0.76 percentage points.

Training on a smaller random subset can therefore outperform training on Full under the same training settings and budget.

\textbf{Observation 2: the effect of dataset size depends on the dataset.}
{
At the same training budget, using fewer images can improve performance on some datasets and reduce it on others. Figure~\ref{fig:empirical_observations_knn} illustrates these differences; results for all seven datasets appear in Appendix~\ref{app:knn_density}. ImageNet is support-limited at high compute: reducing $U$ from Full to roughly half the images already causes a large performance drop, while further halvings continue to hurt but produce smaller drops. On ImageNet, reducing the image pool to roughly half of Full lowers kNN accuracy by 4.65 and 4.56 percentage points at 100M and 400M presentations, respectively. nuPlan shows a tendency similar to ImageNet, with the smallest subsets generally performing worse at higher compute, although the differences are much smaller and the relationship is not strictly monotonic.
In contrast, YFCC15M is less sensitive to dataset size at fixed compute. At $C=400$M, subsets containing at least 25\% of Full achieve similar average performance, ranging from 59.05\% to 59.53\%. At lower budgets, smaller subsets can outperform larger ones: at $C=50$M, the 6.25\% subset exceeds Full by 1.44 percentage points.

Waymo is much more repetition-tolerant: after excluding runs with known training instabilities, reducing $U$ changes performance only weakly at fixed compute. Maxar, SATLAS, and Ego4D reverse the usual intuition: substantially reduced support can outperform Full at fixed compute.

Under logistic regression at 200M, Full also performs best on ImageNet, while smaller subsets outperform it on Maxar, SATLAS, Ego4D, and YFCC15M.

Thus, the fixed-compute trade-off is not a universal curve: reducing unique support can substantially hurt, have a small effect, or improve downstream performance.
}

\textbf{Observation 3: the trade-off changes with compute.}
{
Within a dataset, the performance trade-off between $U$ and $R$ changes with training budget $C$. On ImageNet, the half-size subset outperforms Full by 0.93 percentage points at 20M presentations, but falls 4.65 points below Full at 100M. On Ego4D, Full is the best evaluated subset at 50M presentations, whereas the 6.25\% subset leads at 100M and 200M. YFCC15M shows a different progression: the best evaluated subset changes from 6.25\% at 50M to Full at 100M and 50\% at 200M.

A similar pattern appears on SATLAS, where the 6.25\% subset performs best at 50M and 100M, while the 12.5\% subset leads at 200M. Waymo and nuPlan also show changes in subset ranking, although the performance differences are small.

Maxar provides a contrasting, persistent pattern: its random subset with $U=0.015F$ is the best evaluated subset at every measured budget from 50M to 400M presentations.

Rankings also change under logistic regression: ImageNet's best evaluated subset shifts from 25\% of Full at 20M to Full at 100M.

Thus, the subset that performs best at a low budget may not remain best at a higher budget. 
}

\textbf{Observation 4: repetition can help, saturate, or hurt.}
At fixed support, additional presentations increase repetition once the first pass is complete. Small Maxar subsets improve strongly at lower budgets and then largely saturate. Full Maxar loses 1.63 percentage points of kNN accuracy between 50M and 400M presentations. ImageNet Full initially improves from 65.27\% at 50M to 67.43\% at 100M, then declines to 66.26\% at 400M. In contrast, Ego4D's $U\approx0.0625F$ subset improves from 49.69\% at 50M to 51.26\% at 400M. Additional exposure therefore has different effects depending on the dataset and support size. The same range of outcomes---improvement, saturation, and decline---also appears in the logistic-regression experiments, although individual trends can differ from kNN.

{Together, these observations motivate a density-aware view of image self-distillation. The empirical object is not compute alone, and not unique data alone, but the interaction among total compute $C$, unique support $U$, and repetition density $R$. Existing language-model heuristics are useful baselines, but their assumptions about repeated data do not consistently transfer to image self-distillation.}
\section{Further Analysis}
\label{sec:further_analysis}

We next examine whether the observed subset advantage persists across model sizes, self-distillation methods, subset-selection strategies, and augmentation settings.

\paragraph{Effect of model size.}
We extend the Maxar experiments to iBOT with ViT-S and ViT-L, evaluating four dataset sizes at budgets from 10M to 200M image presentations. At 100M presentations, the random 0.75M-image subset outperforms Full by 2.12 and 4.00 percentage points under kNN for ViT-S and ViT-L, respectively. The corresponding gains under logistic regression are 1.01 and 1.25 percentage points. Together with the ViT-B results, this shows that the subset advantage occurs across all three tested model sizes. Further results appear in Appendix~\ref{app:model_method_density}.

\paragraph{Effect of the self-distillation method.}
The subset advantage also appears with DINOv2, which we train with ViT-B on Maxar. At 100M presentations, the random 0.75M-image subset outperforms Full by 0.87 percentage points under kNN and also performs better under logistic regression. The kNN advantage persists through 300M presentations. These results show that the smaller-subset advantage occurs with both iBOT and DINOv2. Further results appear in Appendix~\ref{app:model_method_density}.

\paragraph{Effect of subset selection.}
\label{subset_choice}
On Maxar, we compare random sampling, hierarchical selection, DASSOT, and NIQE-based quality selection at the same subset size and training budgets. At 50M image presentations, the 0.75M-image subsets achieve average kNN accuracies of 92.02\%, 91.87\%, 92.02\%, and 92.16\%, respectively, compared with 90.15\% for full-data training. All four methods also outperform Full under logistic regression at 50M presentations. At 100M, however, hierarchical selection and DASSOT fall slightly below Full, while random and NIQE selection remain above it (Appendix~\ref{app:subset_selection}). Thus, the advantage at 50M does not require a particular selection strategy, but it does not persist for every method at higher budgets.

\paragraph{Effect of augmentations.}
We test whether repeated exposure remains useful when augmentation is reduced. We train iBOT on a random 0.75M-image Maxar subset with reduced augmentation. We disable color jitter, grayscale, Gaussian blur, and solarization, while retaining cropping, horizontal flipping, normalization, and masking. 

Reduced augmentation, as it expected, lowers average kNN accuracy at 100M presentations from 92.51\% to 89.10\%. However, accuracy increases from 84.00\% at 10M presentations to 89.69\% at 400M. Logistic-regression accuracy also improves with continued training under reduced augmentation. Performance continues to improve with training under reduced augmentation. We retain cropping, horizontal flipping, and masking to support stable iBOT pretraining. Because these operations still introduce variation, this experiment does not establish whether the benefit of repetition is independent of augmentation. The kNN compute curves and logistic-regression comparisons appear in Appendix~\ref{app:reduced_augmentation_density}.

\section{Density-Aware Scaling Law}
\label{scaling_law}
The empirical observations above raise a quantitative question: can existing scaling-law parameterizations explain the interaction among total compute, unique data, and repetition in image self-distillation?

\paragraph{Fitting protocol.}
We fit every law independently for each pretraining dataset. The target is average downstream accuracy across the corresponding frozen-representation evaluation tasks, so the reported error measures how well a law predicts transfer performance rather than pretraining loss. Unless stated otherwise, the main results use the kNN aggregate. Parameters are optimized only on low-compute runs with $C \leq 50$M image views and evaluated on higher-compute runs with $C>50$M. All laws use $C$, $U$, $U_{\mathrm{seen}}$, and $R$ as defined in Section~\ref{sec:sampling_strategies}. We measure \(C\) and \(U_{\mathrm{seen}}\) in millions of image presentations and unique images, respectively. Repetition \(R\) is dimensionless. We apply the run exclusions listed in Appendix~\ref{app:run_exclusions}. We report mean absolute error (MAE) separately in the fit range and in the extrapolation range. Confidence intervals for held-out MAE are reported in Appendix~\ref{app:scaling_law_confidence}. The fitting objective, parameter bounds, initialization procedure, and optimization settings are detailed in Appendix~\ref{app:scaling_fit_details}.

\begin{figure*}[t]
    \centering
    \includegraphics[width=\textwidth]{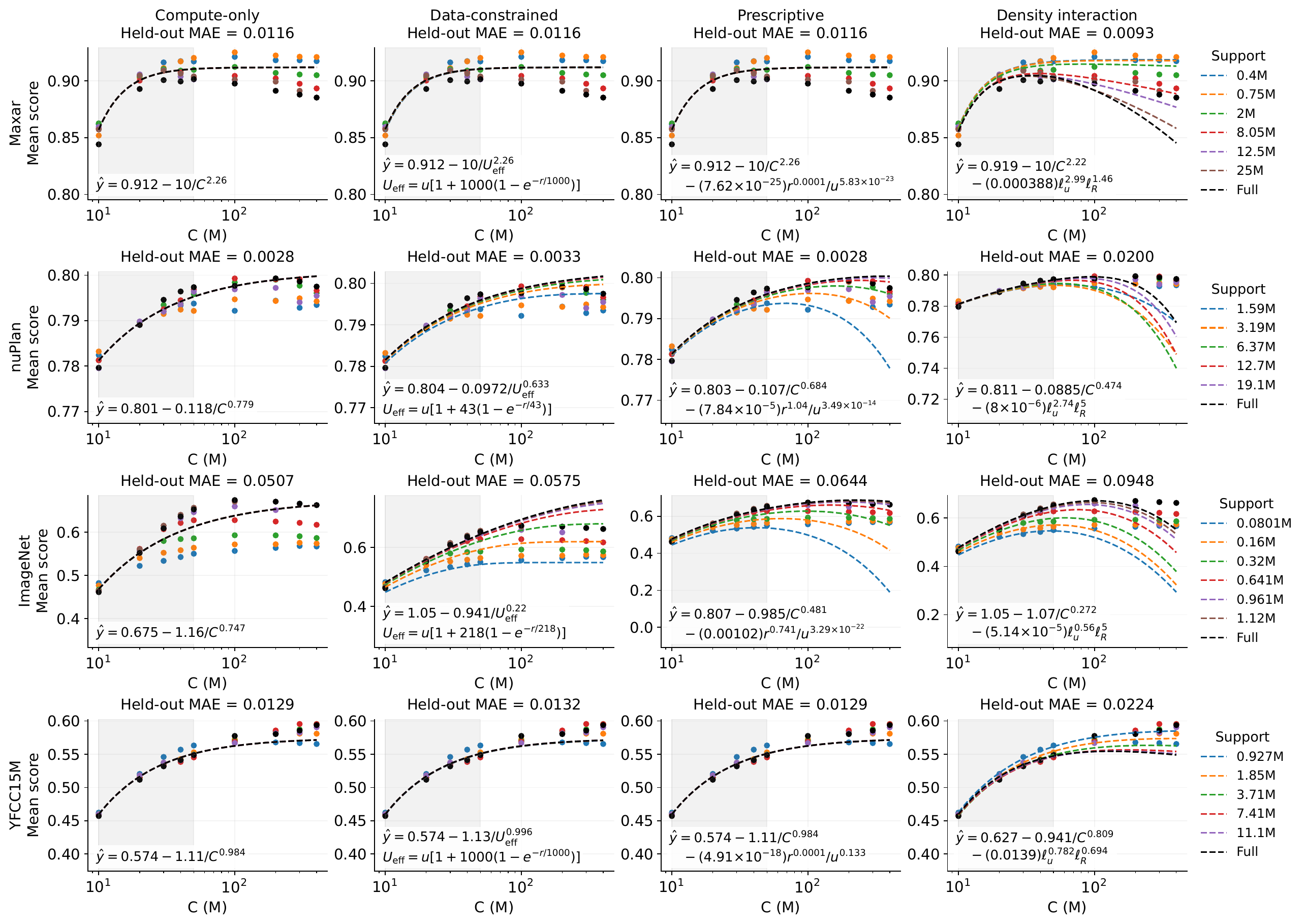}
    \caption{
    Scaling-law fits and extrapolations under the kNN evaluation protocol. Columns compare three existing-law baselines; compute-only, data-constrained, and prescriptive with our density-interaction law. These four forms test whether compute alone, diminishing returns from repetition, a repetition penalty, or a support--repetition interaction captures the observed behavior. Rows show Maxar, nuPlan, ImageNet, and YFCC15M. For compactness, SATLAS and Ego4D are omitted because they show similar qualitative patterns to Maxar, while Waymo shows similar observed trends to nuPlan. Curves are fitted only on the shaded low-compute region and evaluated on higher-compute runs. Fitting curves for all seven datasets are provided in Appendices~\ref{app:knn_scaling} (kNN) and~\ref{app:logres_scaling} (logistic regression).
    }
    \label{fig:scaling_law_main_knn}
\end{figure*}

\paragraph{Existing laws.}
We compare three baseline parameterizations adapted to predict downstream accuracy at fixed model size. The first baseline is a compute-only Chinchilla-style power law{~\cite{chinchillalaw}},
\begin{equation}
    \hat{y}(C)=L-\frac{A}{C^\alpha},
    \label{eq:compute_only}
\end{equation}
which captures improvement with total training exposure but cannot distinguish two runs with the same $C$ and different $(U,R)$. The second baseline uses a data-constrained effective-data parametrization\textcolor{green!50!black}{~\cite{huggingfacelaw}}:
\begin{equation}
    U_{\mathrm{eff}}(U_{\mathrm{seen}},R)=U_{\mathrm{seen}}[1+\tau\left(1-\exp\left(-\frac{R-1}{\tau}\right)\right)],
    \qquad
    \hat{y}(U,R)=L-\frac{A}{U_{\mathrm{eff}}(U_{\mathrm{seen}},R)^\beta}.
    \label{eq:data_constrained}
\end{equation}

{Additional passes provide diminishing benefits, with $\tau$ controlling their saturation.} 
The third baseline is an adaptation of the prescriptive law{~\cite{lovelace2026prescriptive}}, which adds a repetition-dependent penalty: %
\begin{equation}
    \hat{y}(C,U,R)=L-\frac{A}{C^\alpha}-\frac{B(R-1)^{\delta}}{U_{\mathrm{seen}}^{\eta}}.
    \label{eq:prescriptive}
\end{equation}

The compute-only law gives the same prediction to all subsets at a fixed compute budget. The data-constrained law allows repeated exposure to help, but cannot predict declining performance at fixed unique support. The prescriptive law can model such a decline. However, under our parameter constraints, none of these baselines can predict that a smaller subset strictly outperforms a larger one at the same compute budget. Their fitted predictions are also similar on several datasets; Appendix~\ref{app:scaling_law_comparison} explains these similarities and reports the baseline parameters.

\paragraph{Density interaction.}
To capture these differences, we extend the compute-only baseline with a joint correction for unique support and repetition.
\begin{equation}
    \hat{y}(C,U,R)
    = L-\frac{A}{C^\alpha}
    -D[\log(1+U_{\mathrm{seen}})]^\beta[\log(1+R)]^\gamma .
    \label{eq:density_interaction}
\end{equation}
The compute term captures improvement with total training exposure. 
The interaction term adjusts the prediction according to how this exposure is divided between unique images and repetitions. We interpret it as a phenomenological density penalty, not a first-principles law. One possible explanation is that smaller random subsets omit some images that are unhelpful for the downstream tasks. However, subsampling also changes coverage and repetition, and our experiments do not separate these effects. The logarithms avoid treating nominal unique images or repetitions as linearly independent units, and the exponents $\beta$ and $\gamma$ control dataset-specific sensitivity to unique support and repetition density.

\begin{table}[t]
\centering\small
\setlength{\tabcolsep}{2pt}
\renewcommand{\arraystretch}{1.2}
\caption{kNN scaling-law MAE in accuracy units, fitted on $C\leq50$M and evaluated on $C>50$M. Only eligible random subsets and Full are included. Bold and underlining indicate the lowest and second-lowest distinct unrounded held-out errors in each dataset.}
\vspace{6pt}
\label{tab:scaling_law_knn}
\resizebox{\linewidth}{!}{%
\begin{tabular}{lrrrrrrrrrrrrrr}
\toprule
Law & \multicolumn{2}{c}{Maxar} & \multicolumn{2}{c}{SATLAS} & \multicolumn{2}{c}{Waymo} & \multicolumn{2}{c}{nuPlan} & \multicolumn{2}{c}{Ego4D} & \multicolumn{2}{c}{ImageNet} & \multicolumn{2}{c}{YFCC15M} \\
\cmidrule(lr){2-3}\cmidrule(lr){4-5}\cmidrule(lr){6-7}\cmidrule(lr){8-9}\cmidrule(lr){10-11}\cmidrule(lr){12-13}\cmidrule(lr){14-15}
 & fit & extrap. & fit & extrap. & fit & extrap. & fit & extrap. & fit & extrap. & fit & extrap. & fit & extrap. \\
\midrule
Compute-only & 0.0050 & 0.0116 & 0.0022 & 0.0060 & 0.0019 & \textbf{0.0021} & 0.0010 & \underline{0.0028} & 0.0012 & \underline{0.0033} & 0.0234 & \textbf{0.0507} & 0.0033 & \textbf{0.0129} \\
Data-constrained & 0.0050 & 0.0116 & 0.0022 & 0.0058 & 0.0017 & 0.0024 & 0.0009 & 0.0033 & 0.0012 & 0.0033 & 0.0125 & \underline{0.0575} & 0.0034 & 0.0132 \\
Prescriptive & 0.0050 & \underline{0.0116} & 0.0020 & \underline{0.0057} & 0.0015 & 0.0023 & 0.0009 & \textbf{0.0028} & 0.0011 & 0.0034 & 0.0110 & 0.0644 & 0.0033 & \underline{0.0129} \\
Density interaction & 0.0032 & \textbf{0.0093} & 0.0016 & \textbf{0.0043} & 0.0015 & \underline{0.0022} & 0.0008 & 0.0200 & 0.0012 & \textbf{0.0027} & 0.0080 & 0.0948 & 0.0019 & 0.0224 \\
\bottomrule
\end{tabular}}
\end{table}

\paragraph{Main results.}
Table~\ref{tab:scaling_law_knn} shows that density interaction achieves the lowest held-out MAE on Maxar, SATLAS, and Ego4D, and remains close to the best baseline on Waymo. However, it performs worse on nuPlan, ImageNet, and YFCC15M. Under logistic regression, density interaction achieves the lowest held-out MAE among the four compared laws only on Ego4D; a baseline performs better on each of the other six datasets (Appendix~\ref{app:logres_scaling}). Thus, allowing a smaller-subset advantage improves prediction in some settings but does not guarantee better extrapolation.

\paragraph{Limitations.}
Our main experiments use iBOT ViT-B/16; ViT-S, ViT-L, and DINOv2 checks cover Maxar only. Hyperparameters are fixed across subset sizes, and evaluation uses frozen features. The law is fitted on at most 50M presentations and tested up to 400M; its coefficients are descriptive, not universal. Our experiments do not separate the effects of using different images and repeating them more often. All training runs use a single seed, leaving robustness across training seeds untested.

\section*{Conclusion}
\label{conclusion}
Our experiments show that using more unique images does not always improve image self-distillation at a fixed training budget. Smaller random subsets can perform better, but the outcome depends on the dataset, training budget, and evaluation protocol. Additional training can improve performance, lead to a plateau, or reduce performance. The density-interaction law captures these varied patterns, although its predictions do not consistently outperform the adapted baselines.

\begin{ack}
We thank Evan Shelhamer for useful comments. This research was supported by the HESC MESCS RA grant No. 24RL-1B049.
\end{ack}

\bibliographystyle{nips}
\bibliography{ref}

\appendix
\begingroup

\section{Construction of pretraining collections}
\label{app:pretraining_collections}

We construct the Maxar collection from satellite imagery provided through the Maxar Open Data Program~\citep{maxar}. We extract square crops with side lengths randomly chosen from $\{350,400,450,500\}$ pixels. Black image regions are excluded when constructing the collection. For SATLAS we use Sentinel-2 image tiles from SatlasPretrain~\citep{satlas}.

For Waymo, we use camera images from the Waymo Open Dataset End-to-End Driving v1.0.0 release~\citep{waymo_e2e}. For nuPlan, we use camera images distributed in the v1.1 training sensor archives~\citep{nuplan} and construct the pretraining collection from the extracted image files.

For Ego4D~\citep{ego4d}, we extract frames from the \texttt{video\_540ss} videos using non-overlapping windows of 10 frames. We select the first non-black frame in each window, searching forward when necessary and avoiding repeated insertion of the same frame. And we use the standard ImageNet-1K training split~\citep{imagenet,ilsvrc}, containing 1,281,167 images across 1,000 classes.

Our YFCC15M collection is derived from the OpenAI-curated YFCC100M subset released with CLIP~\citep{yfcc100m,clip,openai_yfcc_subset}. Of the 14,829,396 images listed in the released subset, 4,160 were unavailable for download from the source, leaving 14,825,236 images for self-supervised pretraining.
\par
\endgroup

\begingroup

\section{DASSOT selection details}
\label{app:dassot_selection}

For the 0.75M-image Maxar comparison, we adapt DASSOT~\citep{van2024graph} by partitioning pretrained DINOv2 [CLS] embeddings~\citep{dinov2} into 50,000 $k$-means clusters. For a target subset of size $U$, the unrounded quota for cluster $i$ is
\[
q_i^* = U\frac{n_i^{0.7}}{\sum_j n_j^{0.7}},
\]
where $n_i$ is the number of images in that cluster. Quotas are rounded down, capped by cluster size, and the remaining allocation is distributed according to fractional remainders among clusters with available images until the total equals $U$.

Within each cluster, we use a greedy approximation initialized with the image nearest the centroid. At each subsequent step, we select the remaining image with the lowest mean cosine similarity to the already selected images, using normalized embeddings. We combine the selected images across clusters to obtain the final subset. This adaptation serves as a semantic deduplication procedure based on cosine similarity. The 0.75M-image subset contains 750,000 distinct selected images.
\par
\endgroup

\begingroup

\section{Pretraining schedules}
\label{app:training_schedule}

We use a warmup--stable--decay (WSD) learning-rate schedule~\citep{wsd}, with linear decay in our implementation. At global batch size $B=1600$, the learning rate increases linearly from zero to its dataset-specific peak over the first 625 optimizer steps, remains constant until 90\% of the target training budget, and decreases linearly to $2\times10^{-6}$ over the final 10\%. The configured base learning rate is scaled by $B/256$ to obtain the peak learning rate; the final learning rate is not scaled.

Weight decay remains at $0.04$ before increasing linearly to $0.4$ during the final 10\% of training. Teacher EMA momentum similarly remains at $0.996$ before increasing linearly to $1$ during the final 10\%. These schedules use the target budget of the corresponding training endpoint.

Budget-specific endpoints share a constant-learning-rate training trajectory where applicable. For each target budget, training branches from the corresponding checkpoint and completes its final decay phase. Loading a checkpoint restores the student, teacher, optimizer, mixed-precision scaler, and iBOT loss state. Thus, endpoints at different budgets may share a training prefix while completing separate decay phases.
\par
\endgroup

\section{Compute resources}
\label{app:compute_resources}

The main iBOT ViT-B/16 experiments use eight NVIDIA H100 GPUs, each with 80~GB of HBM3 memory, in distributed training. We use 200 images per GPU and a global batch size of 1,600. At this batch size, training budgets of 50M, 100M, 200M, and 400M image presentations correspond to 31,250, 62,500, 125,000, and 250,000 optimizer steps, respectively. Budget-specific endpoints may share a training prefix, as described in Appendix~\ref{app:training_schedule}.

Processing a single $224\times224$ input is estimated to require approximately 241~GFLOPs for the forward and backward passes combined. This estimate is per input; computing the full training cost also requires accounting for all crops and teacher forward passes.

\begingroup
\section{Hyperparameter search}
\label{app:optuna_search}

We tune pretraining hyperparameters separately for Maxar, ImageNet, Waymo, SATLAS, Ego4D, nuPlan, and YFCC15M using Optuna's tree-structured Parzen estimator (TPE) sampler~\citep{optuna}. We search the learning rate, last-layer normalization, maximum masking ratio, maximum local-crop scale, and teacher temperatures for class and patch tokens. Trials are ranked by their final logged iBOT training loss, and the chosen settings are kept fixed across subset sizes within each dataset.

Learning rates are sampled logarithmically, while the other settings are selected from discrete choices. Last-layer normalization is either enabled or disabled, and each teacher temperature is chosen from $\{0.04,0.05,0.07\}$. For ImageNet and Waymo, the maximum masking ratio is chosen from $\{0.3,0.5,0.7\}$ and the maximum local-crop scale from $\{0.25,0.32\}$. For the other four datasets, these choices are $\{0.5,0.7,0.9\}$ and $\{0.25,0.32,0.45\}$, respectively.
\par
\endgroup

\section{Downstream tasks}
\label{app:downstream_tasks}
\FloatBarrier

Maxar and SATLAS are evaluated on six remote-sensing classification tasks (Table~\ref{tab:setup-downstream-maxar}). Waymo and nuPlan use three driving classification tasks, with two additional BDD100K tracking scores included in their kNN aggregates (Table~\ref{tab:setup-downstream-waymo}). Ego4D, ImageNet, and YFCC15M share 24 classification tasks; Ego4D additionally uses EPIC-KITCHENS-100 noun recognition with single-frame and five-frame pooled features (Table~\ref{tab:setup-downstream-imagenet}).

\begingroup

\subsection{Evaluation protocols}
\label{app:downstream_protocol}

\paragraph{Frozen features and kNN.}
Classification uses $\ell_2$-normalized final-layer [CLS] embeddings from the frozen iBOT teacher backbone. We construct a cosine-similarity nearest-neighbor index from the downstream training features and predict each query label by unweighted majority voting among its $k$ nearest neighbors. Voting ties select the smallest class label. For each pretraining group and task, we choose $k\in\{1,5,10,30,50\}$ by mean accuracy of Full-50M. These reference scores use the same evaluation split as the reported task results. The chosen $k$ is fixed across all models and budgets for that group and task, including the added classification tasks.

\paragraph{Logistic regression.}
We fit $\ell_2$-regularized logistic regression on normalized downstream training features using Cyanure~\citep{mairal2019cyanure}, keeping the encoder frozen. The regularization grid is $\lambda\in\{3\times10^{-5},3\times10^{-4},3\times10^{-3},10^{-2},1\}$. We select the regularization value with the highest mean evaluation accuracy across the Full-50M reference models within each pretraining group and apply it to all models in that group.

\paragraph{Object association.}
For Waymo and nuPlan representations, we evaluate BDD100K MOT association on validation videos. Object features are extracted using ROIAlign~\citep{he2017mask} pooling over the frozen backbone's patch-feature maps at annotated bounding boxes and are $\ell_2$-normalized. Each query object is matched to the object with the highest cosine similarity in a later frame of the same video. Instance accuracy measures whether the match has the same track identity; category accuracy measures whether it has the same semantic category. We compute each accuracy over all queries in frame pairs containing objects in both frames, including queries whose track is absent from the candidate frame. We average the per-gap accuracies equally over frame gaps $\{1,2,4,8\}$. The resulting instance and category accuracies each contribute one task score to the kNN aggregate and are excluded from the logistic-regression aggregate.

\par
\endgroup

\begin{table}[htbp]
\centering
\caption{Remote-sensing downstream tasks and fixed reference $k$ values.}
\label{tab:setup-downstream-maxar}
\small
\setlength{\tabcolsep}{4pt}
\begin{tabular}{lrrrrr}
\toprule
Task & Train & Eval & Classes & Maxar $k$ & SATLAS $k$ \\
\midrule
Brick Kiln \cite{brick-kiln} & 58,890 & 7,361 & 2 & 10 & 5 \\
EuroSAT \cite{eurosat} & 16,200 & 5,400 & 10 & 5 & 5 \\
RESISC45 \cite{resisc45} & 18,900 & 6,300 & 45 & 1 & 5 \\
UCMerced \cite{ucm} & 1,260 & 420 & 21 & 1 & 1 \\
PatternNet \cite{patternnet} & 18,240 & 6,080 & 38 & 1 & 5 \\
AID \cite{aid} & 6,000 & 2,000 & 30 & 1 & 1 \\
\bottomrule
\end{tabular}
\end{table}

\begin{table}[htbp]
\centering
\caption{Driving downstream tasks and fixed reference $k$ values. MOT association accuracies enter only the kNN aggregate and are averaged over frame gaps $\{1,2,4,8\}$; the protocol is described in Appendix~\ref{app:downstream_protocol}.}
\label{tab:setup-downstream-waymo}
\small
\setlength{\tabcolsep}{4pt}
\begin{tabular}{lrrrrr}
\toprule
Task & Train & Eval & Classes & Waymo $k$ & nuPlan $k$ \\
\midrule
BDD100K scene \cite{bdd100k} & 69,863 & 10,000 & 7 & 30 & 30 \\
BDD100K weather \cite{bdd100k} & 69,863 & 10,000 & 7 & 30 & 30 \\
ACDC condition \cite{acdc} & 1,600 & 2,000 & 4 & 50 & 30 \\
BDD100K MOT instance & -- & -- & -- & -- & -- \\
BDD100K MOT category & -- & -- & -- & -- & -- \\
\bottomrule
\end{tabular}
\end{table}

\begin{table}[H]
\centering
\caption{General and egocentric downstream tasks. Fixed reference $k$ values are shown for Ego4D (Ego), ImageNet (IN), and YFCC15M (YFCC). A dash indicates that the task is not evaluated for that group.}
\label{tab:setup-downstream-imagenet}
\small
\setlength{\tabcolsep}{3pt}
\begin{tabular}{lrrrrrr}
\toprule
Task & Train & Eval & Classes & Ego $k$ & IN $k$ & YFCC $k$ \\
\midrule
ImageNet-1K \cite{imagenet} & 1,281,167 & 50,000 & 1,000 & 30 & 10 & 30 \\
Stanford Cars \cite{cars} & 8,144 & 8,041 & 196 & 1 & 1 & 1 \\
CIFAR-100 \cite{cifar} & 50,000 & 10,000 & 100 & 10 & 10 & 30 \\
CIFAR-10 \cite{cifar} & 50,000 & 10,000 & 10 & 10 & 10 & 10 \\
CUB-200 \cite{cub} & 5,994 & 5,794 & 200 & 30 & 30 & 30 \\
DTD \cite{dtd} & 1,880 & 1,880 & 47 & 10 & 1 & 10 \\
FER-2013 \cite{fer} & 28,709 & 7,178 & 7 & 1 & 1 & 1 \\
FGVC-Aircraft \cite{aircrafts} & 3,334 & 3,333 & 100 & 1 & 1 & 1 \\
Flowers-102 \cite{flowers} & 1,020 & 1,020 & 102 & 1 & 1 & 1 \\
Food-101 \cite{food101} & 75,750 & 25,250 & 101 & 50 & 30 & 50 \\
GTSRB \cite{gtsrb} & 26,640 & 12,630 & 43 & 50 & 30 & 50 \\
MNIST \cite{mnist} & 60,000 & 10,000 & 10 & 10 & 5 & 10 \\
PCAM \cite{pcam} & 262,144 & 32,768 & 2 & 50 & 50 & 50 \\
Oxford-IIIT Pets \cite{pets} & 3,680 & 3,669 & 37 & 10 & 30 & 30 \\
STL-10 \cite{stl10} & 5,000 & 8,000 & 10 & 10 & 5 & 10 \\
Caltech-101 \cite{caltech101} & 3,030 & 5,647 & 101 & 10 & 10 & 1 \\
RESISC45 \cite{resisc45} & 18,900 & 6,300 & 45 & 1 & 5 & 10 \\
UCMerced \cite{ucm} & 1,260 & 420 & 21 & 1 & 1 & 1 \\
EuroSAT \cite{eurosat} & 16,200 & 5,400 & 10 & 1 & 5 & 5 \\
Brick Kiln \cite{brick-kiln} & 58,890 & 7,361 & 2 & 5 & 5 & 5 \\
\midrule
CORe50 identity \cite{core50} & 119,894 & 44,972 & 50 & 10 & 50 & 50 \\
EgoObjects category \cite{egoobjects} & 447,735 & 164,240 & 638 & 50 & 50 & 50 \\
EgoObjects identity \cite{egoobjects} & 27,690 & 124,956 & 3,510 & 1 & 1 & 1 \\
MIT Indoor-67 \cite{indoor67} & 5,360 & 1,340 & 67 & 30 & 10 & 10 \\
EPIC noun (frame) \cite{epickitchens100} & 67,217 & 9,668 & 300 & 10 & -- & -- \\
EPIC noun (pool) \cite{epickitchens100} & 67,217 & 9,668 & 300 & 5 & -- & -- \\
\bottomrule
\end{tabular}

\vspace{0.5em}
\parbox{\linewidth}{\footnotesize
EgoObjects counts refer to object crops. Category classification uses a 638-label vocabulary, with 352 categories observed in training and 275 in evaluation. Identity classification uses capture-disjoint reference and query crops within the official test subset; Train denotes the reference bank. EPIC-KITCHENS-100 counts refer to action segments, with evaluation on public validation. Its 300-label vocabulary contains 289 labels observed in training; all validation segments are retained.}
\end{table}

\FloatBarrier
\FloatBarrier

\section{Run exclusions}
\label{app:run_exclusions}

The exclusions below apply to both kNN and logistic-regression analyses. Training-loss audits identified instability in the listed SATLAS and Waymo runs. Budgets $C$ are given in millions of image presentations, and subset percentages are relative to the corresponding Full pool.

\begin{table}[H]
\centering
\caption{Runs excluded following training-loss audits.}
\label{tab:excluded_unstable_runs}
\small
\begin{tabular}{llll}
\toprule
Dataset & Selection & Subset size & $C$ (M) \\
\midrule
SATLAS & Random & 25\%, 50\%, 75\% & 300 \\
SATLAS & Random & 6.25\%, 12.5\%, 25\%, 50\%, 75\% & 400 \\
SATLAS & Full & 100\% & 400 \\
Waymo & Random & 6.25\% & 200 \\
Waymo & Full & 100\% & 200 \\
\bottomrule
\end{tabular}
\end{table}

\clearpage
\section{kNN density curves}
\label{app:knn_density}
\begin{figure}[H]
    \centering
    \includegraphics[width=\textwidth,height=0.72\textheight,keepaspectratio]{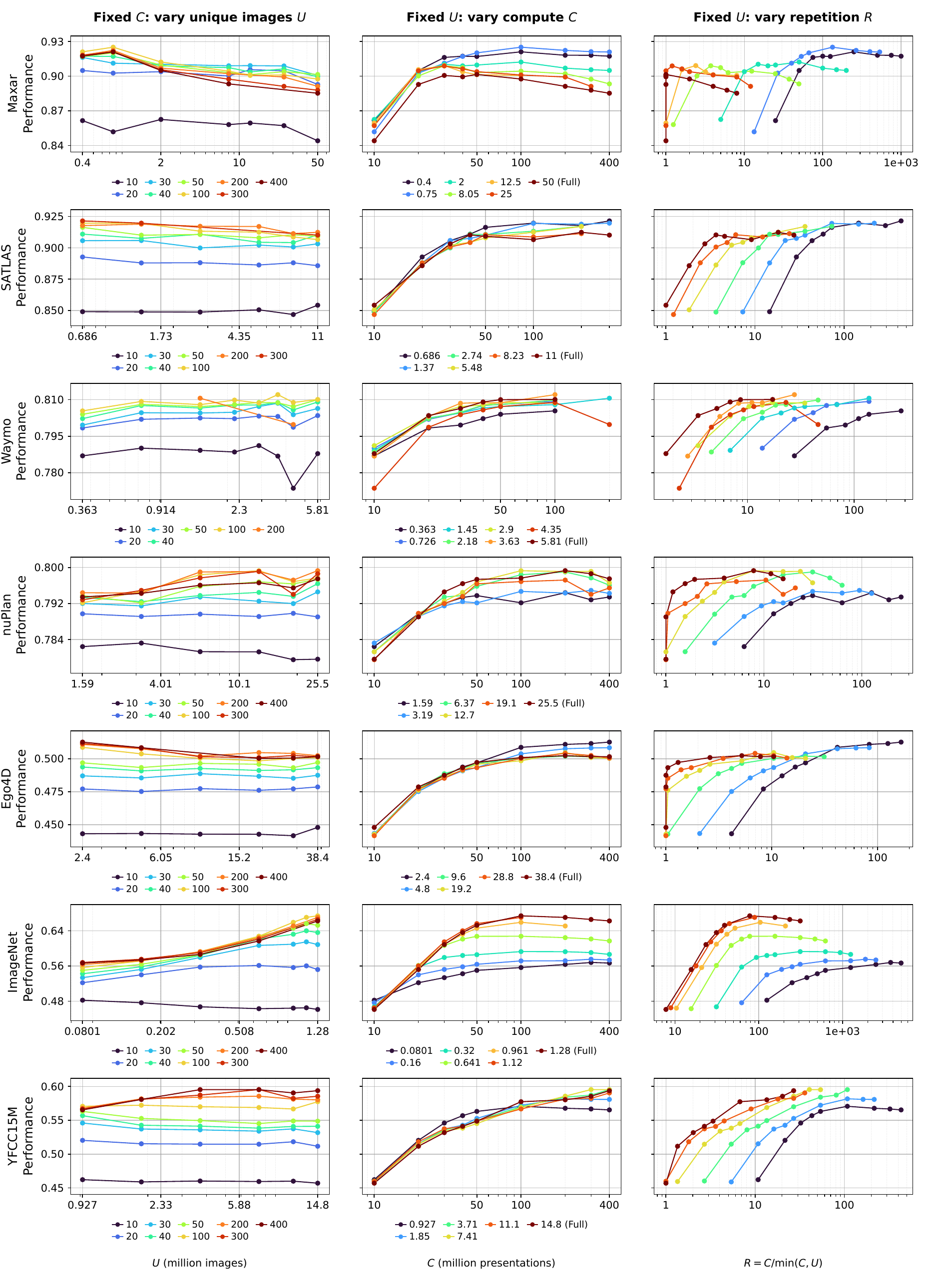}
    \caption{
    Unique-support and repetition trade-offs under kNN evaluation. Rows correspond to Maxar, SATLAS, Waymo, nuPlan, Ego4D, ImageNet, and YFCC15M. The left column fixes compute \(C\) and varies unique images \(U\); the middle column fixes \(U\) and varies \(C\); the right column fixes \(U\) and varies repetition \(R=C/\min(C,U)\). Points show mean downstream accuracy for random-subset and Full-data runs, with lines connecting observations. Colors indicate compute in the left column and unique support in the middle and right columns, which share a color key. For each pretraining group and downstream task, \(k\) is selected from the Full-data 50M reference runs and held fixed across evaluated models. Horizontal axes are logarithmic; image counts are in millions.
    }
    \label{fig:empirical_observations_appdx_knn}
\end{figure}
\clearpage
\section{kNN scaling-law results}
\label{app:knn_scaling}
\begin{figure}[H]
    \centering
    \includegraphics[width=\textwidth,height=0.72\textheight,keepaspectratio]{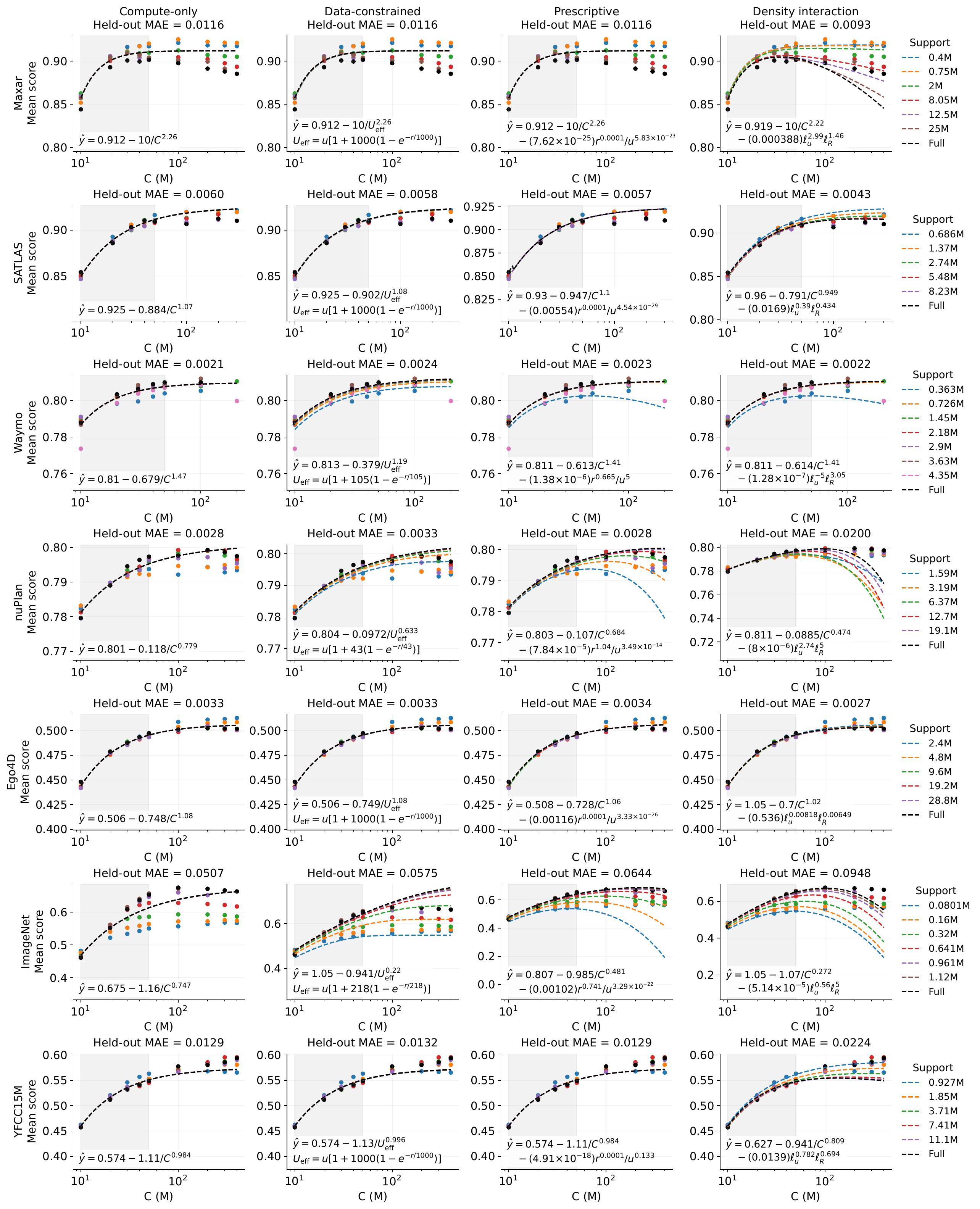}
    \caption{
    Scaling-law fits and extrapolations under the kNN evaluation protocol for all seven datasets. Columns compare the compute-only Chinchilla-style law, the data-constrained effective-data law, the prescriptive scaling law, and our density-interaction law. Rows correspond to Maxar, SATLAS, Waymo, nuPlan, Ego4D, ImageNet, and YFCC15M. Curves are fitted only in the shaded low-compute region ($C\leq50$M) and evaluated on higher-compute runs ($C>50$M). Each panel shows the fitted equation and held-out MAE in accuracy units.
    }
    \label{fig:scaling_law_appdx_knn}
\end{figure}
\clearpage
\section{Logistic-regression density curves}
\label{app:logres_density}

\begin{figure}[H]
    \centering
    \includegraphics[width=\textwidth,height=0.72\textheight,keepaspectratio]{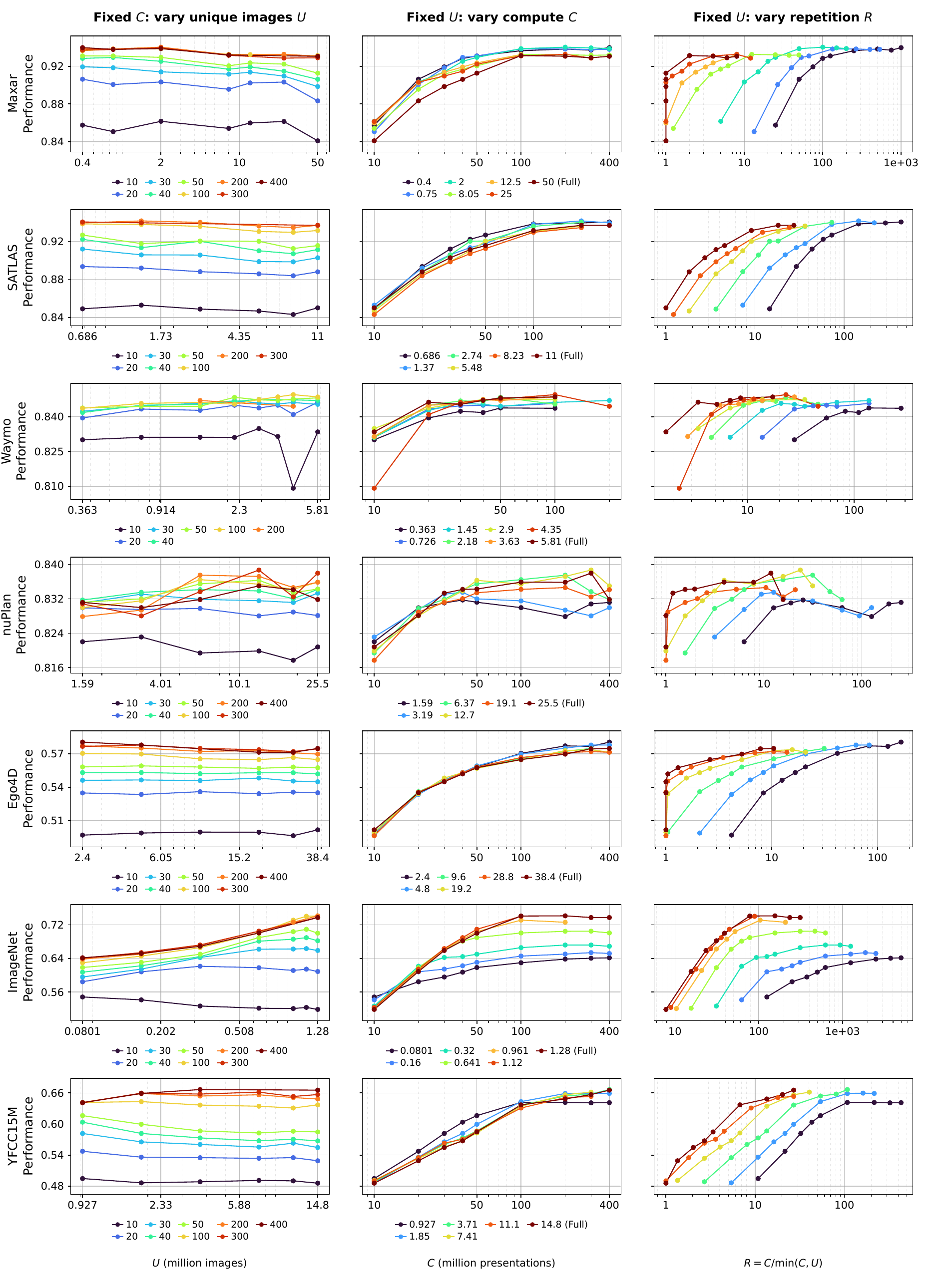}
    \caption{
    Unique-support and repetition trade-offs under logistic-regression evaluation. Rows correspond to Maxar, SATLAS, Waymo, nuPlan, Ego4D, ImageNet, and YFCC15M. Columns show mean downstream accuracy as unique images \(U\) vary at fixed compute \(C\), compute varies at fixed \(U\), and repetition \(R=C/\min(C,U)\) varies at fixed \(U\), respectively. Points represent random-subset and Full-data runs; lines connect observations. Colors indicate compute in the left column and unique support in the middle and right columns, which share a color key. Horizontal axes are logarithmic; image counts are in millions.
    }
    \label{fig:empirical_observations_appdx_logres}
\end{figure}

\clearpage
\section{Logistic-regression scaling-law results}
\label{app:logres_scaling}
\begin{figure}[H]
    \centering
    \includegraphics[width=\textwidth,height=0.72\textheight,keepaspectratio]{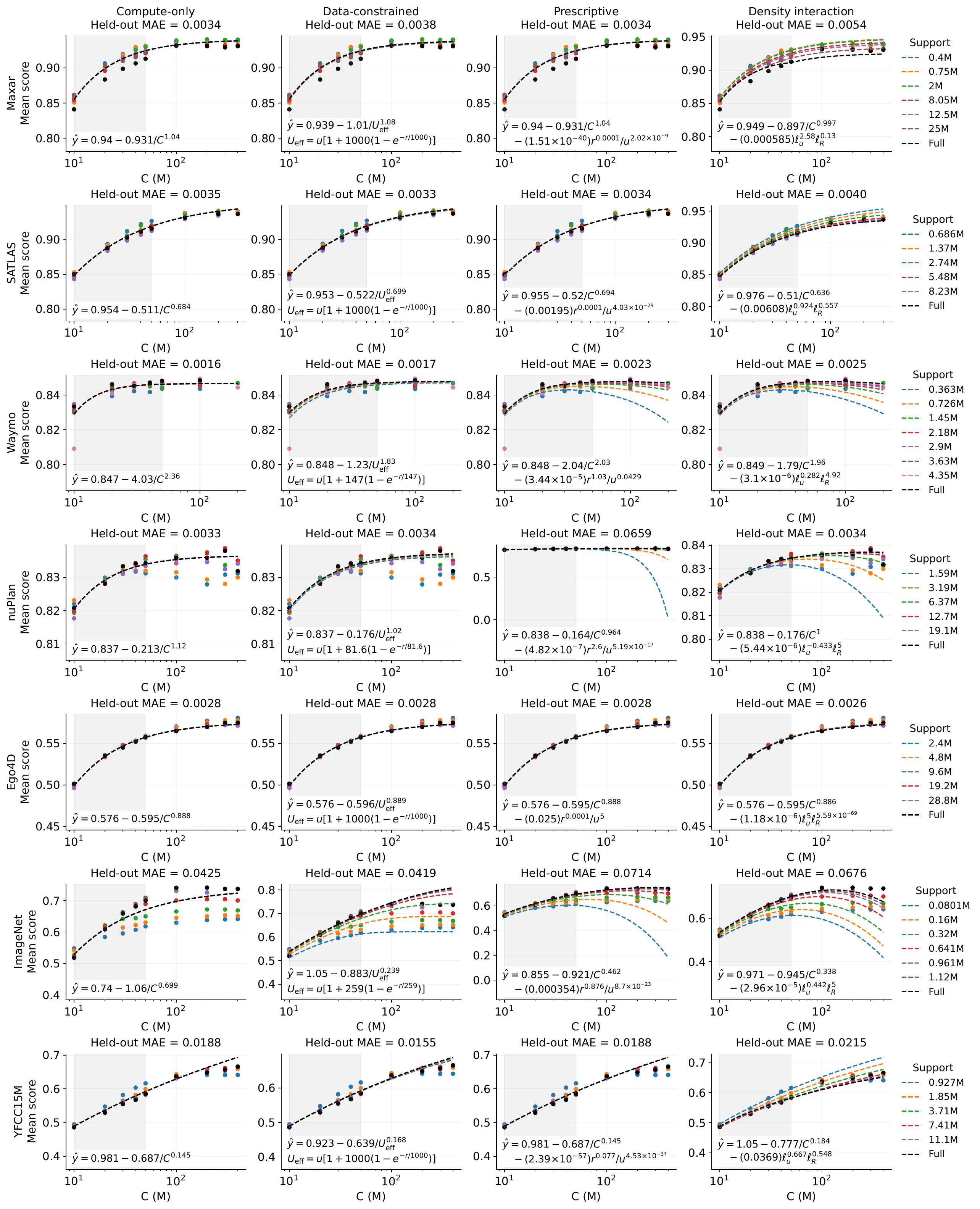}
    \caption{
    Scaling-law fits and extrapolations under the logistic-regression evaluation protocol for all seven datasets. Columns compare the compute-only Chinchilla-style law, the data-constrained effective-data law, the prescriptive scaling law, and our density-interaction law. Rows correspond to Maxar, SATLAS, Waymo, nuPlan, Ego4D, ImageNet, and YFCC15M. Curves are fitted only in the shaded low-compute region ($C\leq50$M) and evaluated on higher-compute runs ($C>50$M). Each panel shows the fitted equation and held-out MAE in accuracy units. Parameters are fitted separately from the kNN evaluation.
    }
    \label{fig:scaling_law_appdx_logres}
\end{figure}

\clearpage
\begin{table}[H]
\centering\small
\setlength{\tabcolsep}{2pt}
\renewcommand{\arraystretch}{1.2}
\caption{Logistic-regression scaling-law MAE for all seven datasets. Each law is fitted on $C\leq50$M and evaluated on $C>50$M, using eligible random subsets and Full. Errors are in accuracy units; lower is better. Parameters are fitted separately from kNN.}
\label{tab:scaling_law_logres}
\resizebox{\linewidth}{!}{%
\begin{tabular}{lrrrrrrrrrrrrrr}
\toprule
Law & \multicolumn{2}{c}{Maxar} & \multicolumn{2}{c}{SATLAS} & \multicolumn{2}{c}{Waymo} & \multicolumn{2}{c}{nuPlan} & \multicolumn{2}{c}{Ego4D} & \multicolumn{2}{c}{ImageNet} & \multicolumn{2}{c}{YFCC15M} \\
\cmidrule(lr){2-3}\cmidrule(lr){4-5}\cmidrule(lr){6-7}\cmidrule(lr){8-9}\cmidrule(lr){10-11}\cmidrule(lr){12-13}\cmidrule(lr){14-15}
 & fit & extrap. & fit & extrap. & fit & extrap. & fit & extrap. & fit & extrap. & fit & extrap. & fit & extrap. \\
\midrule
Compute-only & 0.0056 & 0.0034 & 0.0034 & 0.0035 & 0.0021 & 0.0016 & 0.0011 & 0.0033 & 0.0010 & 0.0028 & 0.0202 & 0.0425 & 0.0068 & 0.0188 \\
Data-constrained & 0.0057 & 0.0038 & 0.0036 & 0.0033 & 0.0019 & 0.0017 & 0.0011 & 0.0034 & 0.0009 & 0.0028 & 0.0112 & 0.0419 & 0.0070 & 0.0155 \\
Prescriptive & 0.0056 & 0.0034 & 0.0034 & 0.0034 & 0.0016 & 0.0023 & 0.0009 & 0.0659 & 0.0010 & 0.0028 & 0.0103 & 0.0714 & 0.0068 & 0.0188 \\
Density interaction & 0.0035 & 0.0054 & 0.0021 & 0.0040 & 0.0016 & 0.0025 & 0.0010 & 0.0034 & 0.0010 & 0.0026 & 0.0090 & 0.0676 & 0.0028 & 0.0215 \\
\bottomrule
\end{tabular}}
\end{table}

\FloatBarrier
\section{Subset selection}
\label{app:subset_selection}
\begin{figure}[h]
    \centering
    \includegraphics[width=0.90\linewidth]{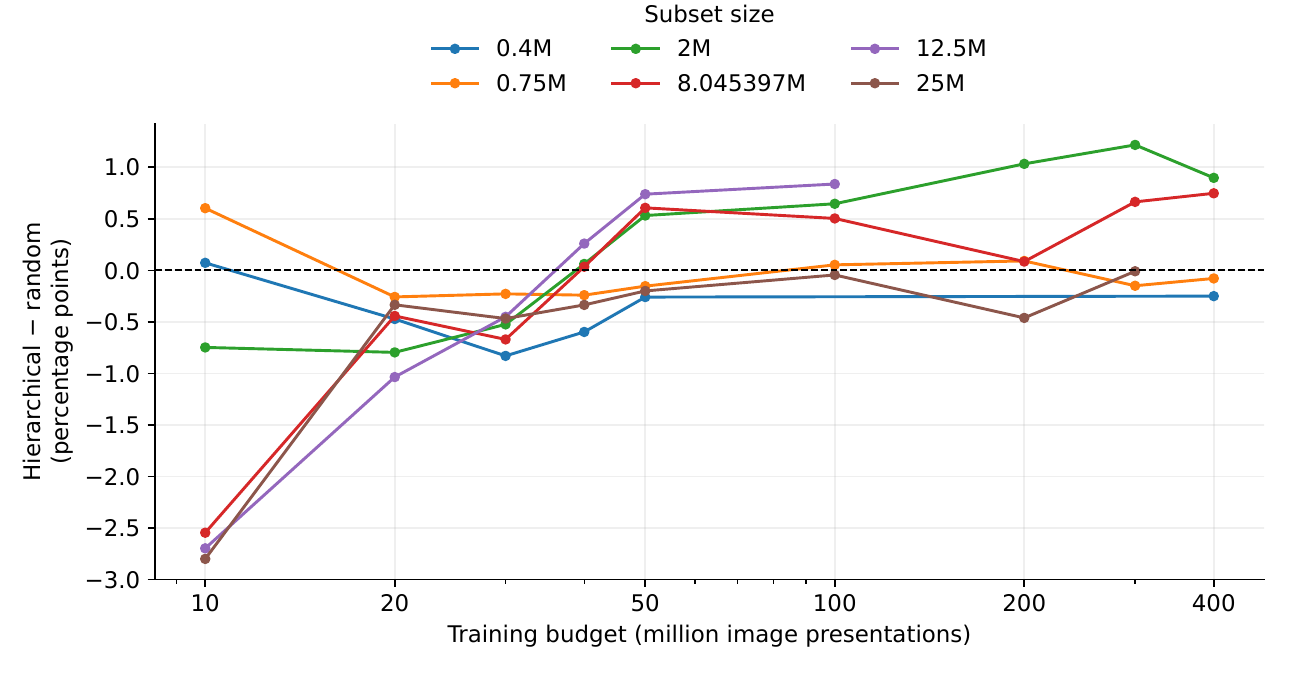}
    \caption{
    Hierarchical versus random subset selection on Maxar under the fixed-reference kNN protocol. Each point compares the same subset size and training budget; values are the difference in averaged downstream scores (hierarchical minus random), expressed in percentage points. Positive values favor hierarchical selection. Lines connect available matched observations for the six reduced subset sizes used in the main analysis. The horizontal axis is logarithmic.}
    \label{fig:maxar_hierarchical_random_lines}
\end{figure}

\begin{table}[htbp]
\centering\small
\setlength{\tabcolsep}{3pt}
\caption{Average downstream kNN and logistic-regression accuracy (\%) for iBOT ViT-B on Maxar. Selected subsets contain 0.75M images; Full contains 50M images. Columns give the training budget in millions of image presentations. Dashes indicate unavailable results.}
\label{tab:subset_selection_logres}
\begin{tabular*}{\linewidth}{@{\extracolsep{\fill}}l*{9}{r}@{}}
\toprule
Selection & 10M & 20M & 30M & 40M & 50M & 100M & 200M & 300M & 400M \\
\midrule
\multicolumn{10}{l}{\textbf{kNN}} \\
\addlinespace[2pt]
Full & 84.41 & 89.28 & 90.07 & 89.95 & 90.15 & 89.76 & 89.12 & 88.79 & 88.53 \\
Random & 85.18 & 90.27 & 91.13 & 91.71 & 92.02 & 92.51 & 92.24 & 92.13 & 92.10 \\
Hierarchical & 85.79 & 90.01 & 90.90 & 91.47 & 91.87 & 92.57 & 92.33 & 91.98 & 92.02 \\
DASSOT & 84.10 & 89.95 & 91.11 & 91.84 & 92.02 & 92.79 & 92.80 & 92.21 & 92.30 \\
NIQE & 86.32 & 89.93 & 91.52 & 91.92 & 92.16 & 92.36 & -- & -- & -- \\
\midrule
\multicolumn{10}{l}{\textbf{Logistic regression}} \\
\addlinespace[2pt]
Full & 84.10 & 88.33 & 89.85 & 90.61 & 91.27 & 93.13 & 93.06 & 92.90 & 93.05 \\
Random & 85.07 & 90.06 & 91.84 & 92.93 & 93.09 & 93.83 & 93.83 & 93.78 & 93.79 \\
Hierarchical & 85.85 & 89.43 & 90.97 & 91.78 & 92.44 & 92.98 & 93.09 & 93.19 & 93.07 \\
DASSOT & 83.49 & 88.37 & 90.32 & 91.56 & 92.09 & 93.08 & 93.19 & 93.26 & 93.32 \\
NIQE & 87.09 & 90.49 & 92.23 & 92.69 & 93.00 & 93.71 & -- & -- & -- \\
\bottomrule
\end{tabular*}
\end{table}

\clearpage
\section{Confidence intervals for scaling-law prediction errors}
\label{app:scaling_law_confidence}

We estimate uncertainty in held-out MAE using a bootstrap over evaluated configurations. Each configuration corresponds to a subset size and training budget, with averaged downstream accuracy. We use the same random-subset and Full runs as in the main scaling analysis, after the exclusions in Appendix~\ref{app:run_exclusions}. The fitted laws remain fixed: fitting uses $C\leq50$M, and the bootstrap uses only configurations with $C>50$M.

For each dataset and evaluation protocol, we sample the same number of held-out configurations with replacement and recompute MAE. We repeat this 10,000 times, using identical sampled configurations for all four laws. Tables~\ref{tab:scaling_law_ci_knn} and~\ref{tab:scaling_law_ci_logres} report the original MAE, bootstrap mean, bootstrap standard deviation (SD), and 95\% percentile interval (the 2.5th and 97.5th percentiles). All errors are in accuracy units.

These intervals describe sensitivity to the composition of the held-out evaluation set. They do not include uncertainty from training seeds or fitting the laws. Configurations are sampled independently.

\begin{table}[H]
\centering\small
\setlength{\tabcolsep}{4pt}
\renewcommand{\arraystretch}{0.95}
\caption{Held-out MAE and bootstrap uncertainty under kNN evaluation. $n$ is the number of held-out configurations; the 95\% intervals use 10,000 resamples with fixed fitted laws.}
\label{tab:scaling_law_ci_knn}
\begin{tabular}{lrrrr}
\toprule
Law & Original MAE & Bootstrap mean & Bootstrap SD & 95\% interval \\
\midrule
\multicolumn{5}{l}{\textbf{Maxar} ($n=24$)} \\
Compute-only & 0.011606 & 0.011612 & 0.001292 & [0.009192, 0.014187] \\
Data-constrained & 0.011625 & 0.011631 & 0.001293 & [0.009212, 0.014209] \\
Prescriptive & 0.011606 & 0.011612 & 0.001292 & [0.009192, 0.014187] \\
Density interaction & 0.009325 & 0.009316 & 0.002087 & [0.005621, 0.013650] \\
\midrule
\multicolumn{5}{l}{\textbf{SATLAS} ($n=15$)} \\
Compute-only & 0.005959 & 0.005953 & 0.000994 & [0.004057, 0.007909] \\
Data-constrained & 0.005828 & 0.005823 & 0.000998 & [0.003931, 0.007781] \\
Prescriptive & 0.005731 & 0.005726 & 0.000979 & [0.003867, 0.007655] \\
Density interaction & 0.004300 & 0.004297 & 0.000648 & [0.003057, 0.005587] \\
\midrule
\multicolumn{5}{l}{\textbf{Waymo} ($n=10$)} \\
Compute-only & 0.002126 & 0.002126 & 0.000874 & [0.000765, 0.004042] \\
Data-constrained & 0.002423 & 0.002422 & 0.001022 & [0.001033, 0.004691] \\
Prescriptive & 0.002271 & 0.002272 & 0.000980 & [0.000728, 0.004438] \\
Density interaction & 0.002208 & 0.002208 & 0.000971 & [0.000699, 0.004365] \\
\midrule
\multicolumn{5}{l}{\textbf{nuPlan} ($n=24$)} \\
Compute-only & 0.002841 & 0.002833 & 0.000440 & [0.001991, 0.003726] \\
Data-constrained & 0.003295 & 0.003288 & 0.000374 & [0.002551, 0.004025] \\
Prescriptive & 0.002792 & 0.002782 & 0.000699 & [0.001583, 0.004298] \\
Density interaction & 0.020019 & 0.020003 & 0.003215 & [0.013920, 0.026362] \\
\midrule
\multicolumn{5}{l}{\textbf{Ego4D} ($n=23$)} \\
Compute-only & 0.003320 & 0.003325 & 0.000460 & [0.002455, 0.004249] \\
Data-constrained & 0.003346 & 0.003351 & 0.000465 & [0.002474, 0.004287] \\
Prescriptive & 0.003391 & 0.003396 & 0.000434 & [0.002561, 0.004248] \\
Density interaction & 0.002733 & 0.002736 & 0.000444 & [0.001904, 0.003638] \\
\midrule
\multicolumn{5}{l}{\textbf{ImageNet} ($n=23$)} \\
Compute-only & 0.050680 & 0.050786 & 0.006496 & [0.038177, 0.063372] \\
Data-constrained & 0.057543 & 0.057397 & 0.006562 & [0.044924, 0.070576] \\
Prescriptive & 0.064431 & 0.064780 & 0.019279 & [0.031269, 0.106046] \\
Density interaction & 0.094760 & 0.094838 & 0.017209 & [0.062180, 0.129468] \\
\midrule
\multicolumn{5}{l}{\textbf{YFCC15M} ($n=24$)} \\
Compute-only & 0.012944 & 0.012949 & 0.001369 & [0.010302, 0.015667] \\
Data-constrained & 0.013203 & 0.013209 & 0.001418 & [0.010429, 0.016002] \\
Prescriptive & 0.012944 & 0.012949 & 0.001369 & [0.010302, 0.015667] \\
Density interaction & 0.022406 & 0.022402 & 0.002461 & [0.017640, 0.027263] \\
\bottomrule
\end{tabular}
\end{table}

\clearpage
\begin{table}[H]
\centering\small
\setlength{\tabcolsep}{4pt}
\renewcommand{\arraystretch}{0.95}
\caption{Held-out MAE and bootstrap uncertainty under logistic-regression evaluation. $n$ is the number of held-out configurations; the 95\% intervals use 10,000 resamples with fixed fitted laws.}
\label{tab:scaling_law_ci_logres}
\begin{tabular}{lrrrr}
\toprule
Law & Original MAE & Bootstrap mean & Bootstrap SD & 95\% interval \\
\midrule
\multicolumn{5}{l}{\textbf{Maxar} ($n=24$)} \\
Compute-only & 0.003445 & 0.003450 & 0.000561 & [0.002371, 0.004571] \\
Data-constrained & 0.003772 & 0.003777 & 0.000468 & [0.002860, 0.004697] \\
Prescriptive & 0.003445 & 0.003450 & 0.000561 & [0.002371, 0.004571] \\
Density interaction & 0.005393 & 0.005390 & 0.000614 & [0.004191, 0.006612] \\
\midrule
\multicolumn{5}{l}{\textbf{SATLAS} ($n=15$)} \\
Compute-only & 0.003536 & 0.003534 & 0.000520 & [0.002530, 0.004564] \\
Data-constrained & 0.003257 & 0.003254 & 0.000569 & [0.002150, 0.004398] \\
Prescriptive & 0.003414 & 0.003412 & 0.000515 & [0.002412, 0.004434] \\
Density interaction & 0.003960 & 0.003963 & 0.000990 & [0.002176, 0.006036] \\
\midrule
\multicolumn{5}{l}{\textbf{Waymo} ($n=10$)} \\
Compute-only & 0.001556 & 0.001560 & 0.000297 & [0.001005, 0.002151] \\
Data-constrained & 0.001668 & 0.001669 & 0.000314 & [0.001081, 0.002316] \\
Prescriptive & 0.002259 & 0.002260 & 0.000616 & [0.001213, 0.003591] \\
Density interaction & 0.002493 & 0.002493 & 0.000648 & [0.001317, 0.003853] \\
\midrule
\multicolumn{5}{l}{\textbf{nuPlan} ($n=24$)} \\
Compute-only & 0.003288 & 0.003287 & 0.000484 & [0.002369, 0.004252] \\
Data-constrained & 0.003434 & 0.003435 & 0.000503 & [0.002474, 0.004430] \\
Prescriptive & 0.065916 & 0.065844 & 0.036143 & [0.010994, 0.147145] \\
Density interaction & 0.003360 & 0.003360 & 0.001037 & [0.001660, 0.005676] \\
\midrule
\multicolumn{5}{l}{\textbf{Ego4D} ($n=23$)} \\
Compute-only & 0.002791 & 0.002793 & 0.000449 & [0.001936, 0.003696] \\
Data-constrained & 0.002827 & 0.002829 & 0.000459 & [0.001952, 0.003754] \\
Prescriptive & 0.002825 & 0.002826 & 0.000460 & [0.001950, 0.003759] \\
Density interaction & 0.002647 & 0.002649 & 0.000416 & [0.001857, 0.003490] \\
\midrule
\multicolumn{5}{l}{\textbf{ImageNet} ($n=23$)} \\
Compute-only & 0.042506 & 0.042526 & 0.005061 & [0.032765, 0.052264] \\
Data-constrained & 0.041866 & 0.041790 & 0.005062 & [0.031921, 0.051657] \\
Prescriptive & 0.071371 & 0.071306 & 0.024094 & [0.030066, 0.123805] \\
Density interaction & 0.067559 & 0.067447 & 0.012777 & [0.043737, 0.093967] \\
\midrule
\multicolumn{5}{l}{\textbf{YFCC15M} ($n=22$)} \\
Compute-only & 0.018817 & 0.018806 & 0.002647 & [0.013861, 0.024177] \\
Data-constrained & 0.015467 & 0.015459 & 0.002022 & [0.011597, 0.019529] \\
Prescriptive & 0.018817 & 0.018806 & 0.002647 & [0.013861, 0.024177] \\
Density interaction & 0.021520 & 0.021480 & 0.003949 & [0.014408, 0.030033] \\
\bottomrule
\end{tabular}
\end{table}

\clearpage
\section{Scaling-law ablation plots}
\label{app:scaling_ablations}

The numerical comparisons below cover all seven datasets under both evaluation protocols. Let $\ell_U=\log(1+U_{\mathrm{seen}})$ and $\ell_R=\log(1+R)$, using the quantities defined in Section~\ref{sec:sampling_strategies}. $C$ and $U_{\mathrm{seen}}$ are measured in millions. The variants are:
\begin{align*}
\text{Density interaction:}\quad & L-A/C^\alpha-D\ell_U^\beta\ell_R^\gamma,\\
\text{Unique-only:}\quad & L-A/C^\alpha-D_U\ell_U^\beta,\\
\text{Repeat-only:}\quad & L-A/C^\alpha-D_R\ell_R^\gamma,\\
\text{Separate-additive:}\quad & L-A/C^\alpha-D_U\ell_U^\beta-D_R\ell_R^\gamma,\\
\text{Raw-}U\text{ interaction:}\quad & L-A/C^\alpha-DU_{\mathrm{seen}}^\beta\ell_R^\gamma,\\
\text{Raw-}R\text{ interaction:}\quad & L-A/C^\alpha-D\ell_U^\beta R^\gamma.
\end{align*}

\begin{table}[H]
\centering\small
\setlength{\tabcolsep}{2pt}
\renewcommand{\arraystretch}{1.2}
\caption{kNN ablation MAE for all seven datasets, fitted on $C\leq50$M and evaluated on $C>50$M. Every variant uses the same eligible observations within each dataset. Lower is better.}
\label{tab:ablation_knn_all}
\resizebox{\linewidth}{!}{%
\begin{tabular}{lrrrrrrrrrrrrrr}
\toprule
Law & \multicolumn{2}{c}{Maxar} & \multicolumn{2}{c}{SATLAS} & \multicolumn{2}{c}{Waymo} & \multicolumn{2}{c}{nuPlan} & \multicolumn{2}{c}{Ego4D} & \multicolumn{2}{c}{ImageNet} & \multicolumn{2}{c}{YFCC15M} \\
\cmidrule(lr){2-3}\cmidrule(lr){4-5}\cmidrule(lr){6-7}\cmidrule(lr){8-9}\cmidrule(lr){10-11}\cmidrule(lr){12-13}\cmidrule(lr){14-15}
 & fit & extrap. & fit & extrap. & fit & extrap. & fit & extrap. & fit & extrap. & fit & extrap. & fit & extrap. \\
\midrule
Density interaction & 0.0032 & 0.0093 & 0.0016 & 0.0043 & 0.0015 & 0.0022 & 0.0008 & 0.0200 & 0.0012 & 0.0027 & 0.0080 & 0.0948 & 0.0019 & 0.0224 \\
Unique-only & 0.0035 & 0.0084 & 0.0019 & 0.0058 & 0.0015 & 0.0018 & 0.0009 & 0.0028 & 0.0012 & 0.0033 & 0.0158 & 0.0392 & 0.0025 & 0.0145 \\
Repeat-only & 0.0050 & 0.0116 & 0.0022 & 0.0060 & 0.0016 & 0.0032 & 0.0009 & 0.0016 & 0.0012 & 0.0033 & 0.0106 & 0.0444 & 0.0033 & 0.0129 \\
Separate-additive & 0.0032 & 0.0093 & 0.0016 & 0.0036 & 0.0014 & 0.0043 & 0.0006 & 0.0025 & 0.0012 & 0.0026 & 0.0097 & 0.0386 & 0.0019 & 0.0272 \\
Raw-$U$ interaction & 0.0030 & 0.0127 & 0.0016 & 0.0042 & 0.0015 & 0.0022 & 0.0008 & 0.0272 & 0.0012 & 0.0025 & 0.0083 & 0.0980 & 0.0020 & 0.0193 \\
Raw-$R$ interaction & 0.0030 & 0.0288 & 0.0016 & 0.0232 & 0.0015 & 0.0023 & 0.0008 & 0.1831 & 0.0012 & 0.0046 & 0.0082 & 0.3793 & 0.0019 & 0.0787 \\
\bottomrule
\end{tabular}}
\end{table}

{
\begin{table}[H]
\centering\small
\caption{Average downstream logistic-regression accuracy (\%) on Maxar across model sizes and self-distillation methods. The random subset contains 0.75M images and Full contains 50M images. Both use the same budget $C$, in millions of image presentations.}
\label{tab:model_method_logres}
\begin{tabular}{lrrr}
\toprule
Configuration & $C$ (M) & Random subset & Full \\
\midrule
iBOT, ViT-S & 50 & 92.03 & 91.17 \\
iBOT, ViT-S & 100 & 92.95 & 91.94 \\
iBOT, ViT-S & 200 & 93.21 & 92.05 \\
iBOT, ViT-B & 50 & 93.09 & 91.27 \\
iBOT, ViT-B & 100 & 93.83 & 93.13 \\
iBOT, ViT-B & 200 & 93.83 & 93.06 \\
iBOT, ViT-L & 50 & 94.17 & 92.96 \\
iBOT, ViT-L & 100 & 94.33 & 93.08 \\
iBOT, ViT-L & 200 & 93.95 & 93.27 \\
DINOv2, ViT-B & 50 & 93.09 & 92.21 \\
DINOv2, ViT-B & 100 & 93.70 & 93.34 \\
DINOv2, ViT-B & 200 & 94.37 & 93.73 \\
\bottomrule
\end{tabular}
\end{table}
}

\begin{table}[H]
\centering\small
\setlength{\tabcolsep}{2pt}
\renewcommand{\arraystretch}{1.2}
\caption{Logistic-regression ablation MAE for all seven datasets, fitted on $C\leq50$M and evaluated on $C>50$M. Every variant uses the same eligible observations within each dataset. Lower is better.}
\label{tab:ablation_logres_all}
\resizebox{\linewidth}{!}{%
\begin{tabular}{lrrrrrrrrrrrrrr}
\toprule
Law & \multicolumn{2}{c}{Maxar} & \multicolumn{2}{c}{SATLAS} & \multicolumn{2}{c}{Waymo} & \multicolumn{2}{c}{nuPlan} & \multicolumn{2}{c}{Ego4D} & \multicolumn{2}{c}{ImageNet} & \multicolumn{2}{c}{YFCC15M} \\
\cmidrule(lr){2-3}\cmidrule(lr){4-5}\cmidrule(lr){6-7}\cmidrule(lr){8-9}\cmidrule(lr){10-11}\cmidrule(lr){12-13}\cmidrule(lr){14-15}
 & fit & extrap. & fit & extrap. & fit & extrap. & fit & extrap. & fit & extrap. & fit & extrap. & fit & extrap. \\
\midrule
Density interaction & 0.0035 & 0.0054 & 0.0021 & 0.0040 & 0.0016 & 0.0025 & 0.0010 & 0.0034 & 0.0010 & 0.0026 & 0.0090 & 0.0676 & 0.0028 & 0.0215 \\
Unique-only & 0.0035 & 0.0048 & 0.0022 & 0.0029 & 0.0017 & 0.0012 & 0.0011 & 0.0032 & 0.0010 & 0.0026 & 0.0149 & 0.0300 & 0.0045 & 0.0151 \\
Repeat-only & 0.0056 & 0.0034 & 0.0034 & 0.0035 & 0.0016 & 0.0021 & 0.0010 & 0.0035 & 0.0010 & 0.0028 & 0.0098 & 0.0419 & 0.0068 & 0.0188 \\
Separate-additive & 0.0035 & 0.0053 & 0.0022 & 0.0030 & 0.0016 & 0.0072 & 0.0007 & 0.0046 & 0.0009 & 0.0028 & 0.0087 & 0.0484 & 0.0039 & 0.0160 \\
Raw-$U$ interaction & 0.0034 & 0.0062 & 0.0021 & 0.0037 & 0.0016 & 0.0024 & 0.0010 & 0.0034 & 0.0009 & 0.0028 & 0.0091 & 0.0682 & 0.0032 & 0.0225 \\
Raw-$R$ interaction & 0.0034 & 0.0063 & 0.0021 & 0.0055 & 0.0016 & 0.0024 & 0.0009 & 2.4109 & 0.0010 & 0.0026 & 0.0085 & 0.4020 & 0.0036 & 0.0375 \\
\bottomrule
\end{tabular}}
\end{table}

\begin{figure}[H]
    \centering
    \includegraphics[width=0.98\textwidth,height=0.72\textheight,keepaspectratio]{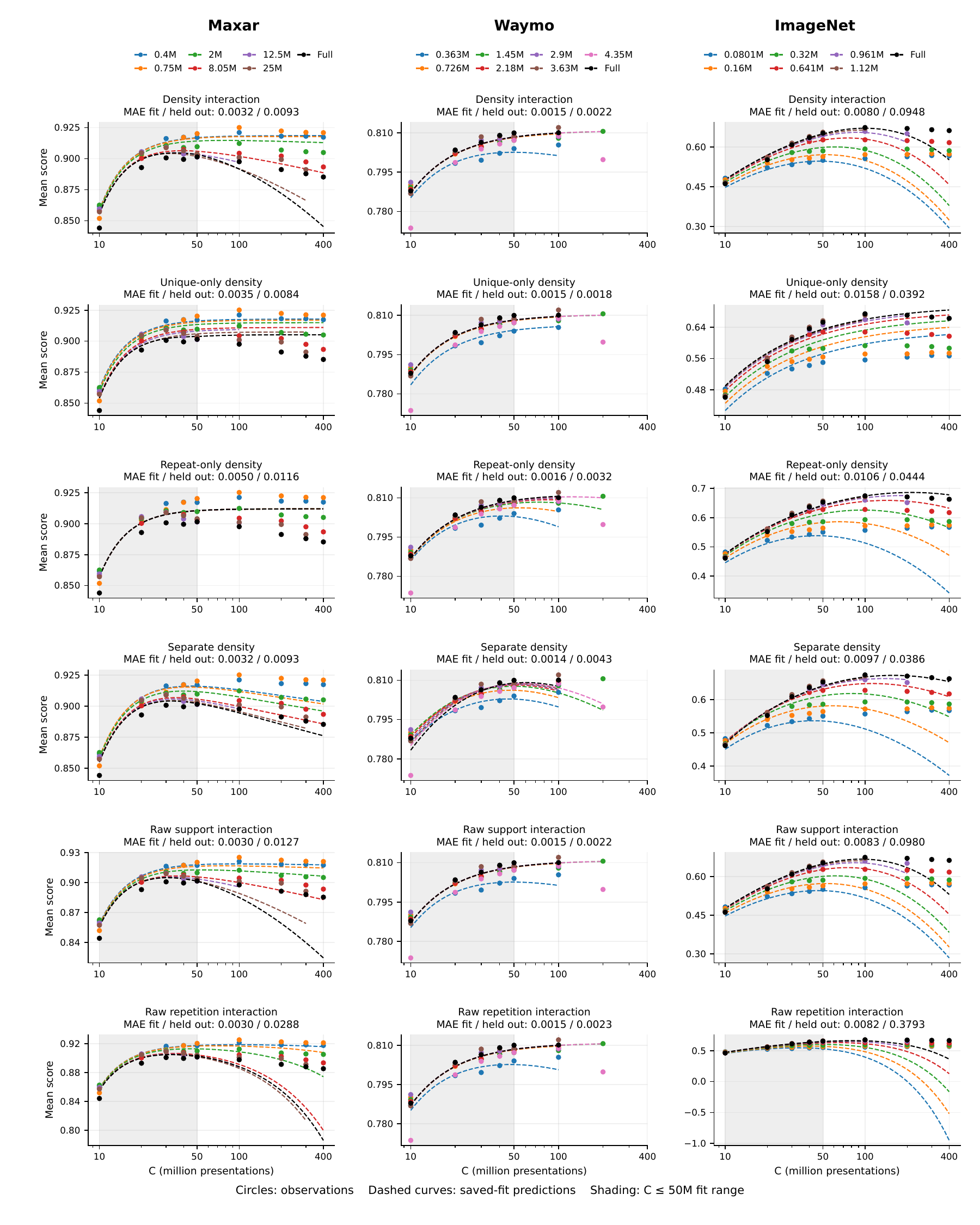}
    \caption{
    kNN ablation plots for the density-interaction law. Columns show Maxar, Waymo, and ImageNet; rows compare the interaction, unique-only, repeat-only, separate-additive, raw-$U_{\mathrm{seen}}$, and raw-$R$ forms. Points show observed scores and dashed curves show predictions from the saved fits. The shaded region denotes the fit range ($C\leq50$M); higher-compute observations are held out. Panel titles report fit and held-out MAE in accuracy units. Support sizes and $C$ are in millions.}
    \label{fig:scaling_law_ablations_knn}
\end{figure}

\begin{figure}[H]
    \centering
    \includegraphics[width=0.98\textwidth,height=0.72\textheight,keepaspectratio]{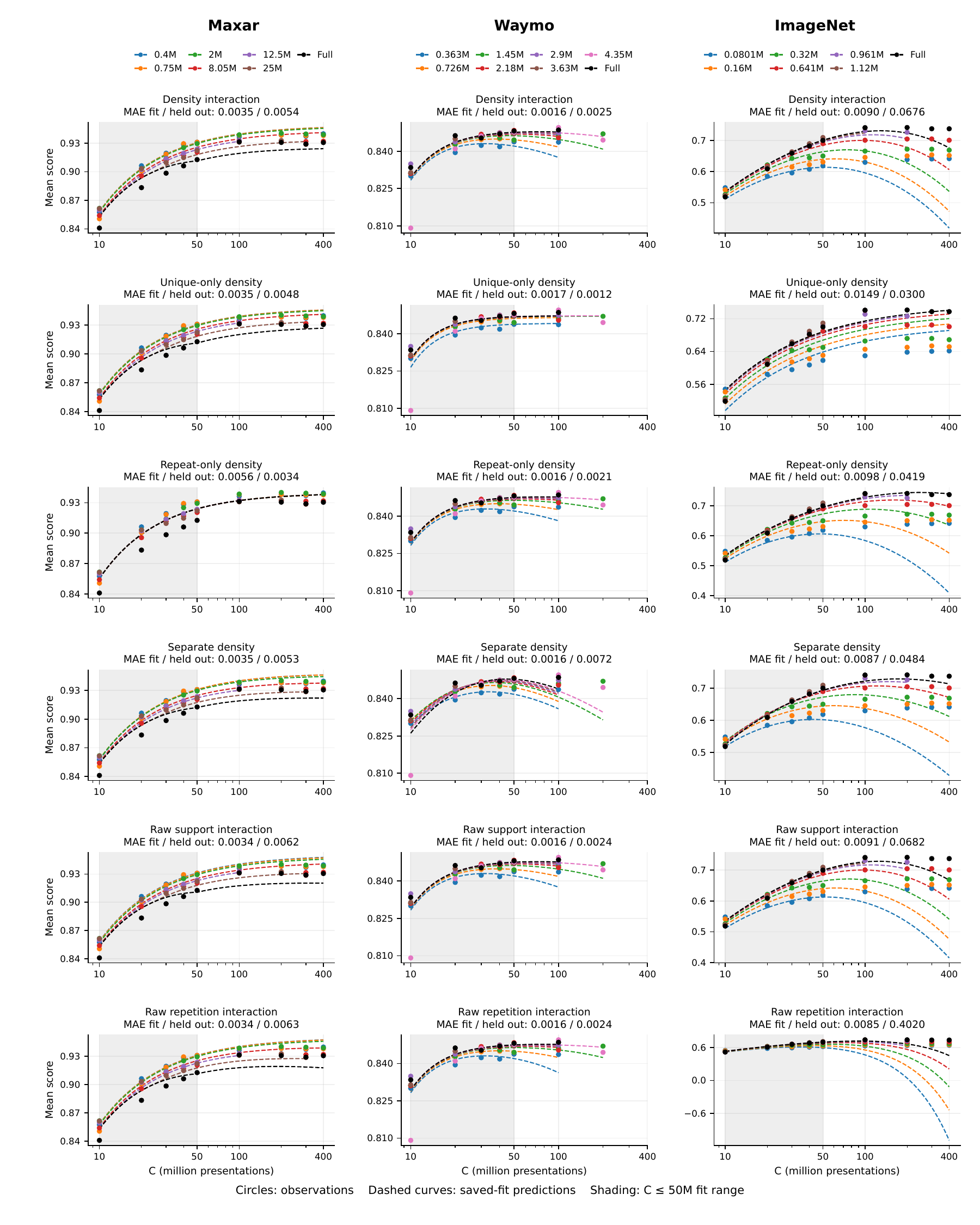}
    \caption{
    Logistic-regression ablation plots for the density-interaction law. Columns show Maxar, Waymo, and ImageNet; rows compare the interaction, unique-only, repeat-only, separate-additive, raw-$U_{\mathrm{seen}}$, and raw-$R$ forms. Points show observed scores and dashed curves show predictions from the saved fits. The shaded region denotes the fit range ($C\leq50$M); higher-compute observations are held out. Panel titles report fit and held-out MAE in accuracy units. Support sizes and $C$ are in millions.}
    \label{fig:scaling_law_ablations_logres}
\end{figure}

Tables~\ref{tab:ablation_knn_all} and~\ref{tab:ablation_logres_all} provide the complete numerical comparisons. For compactness, Figures~\ref{fig:scaling_law_ablations_knn} and~\ref{fig:scaling_law_ablations_logres} show curves only for Maxar, Waymo, and ImageNet; the tables include all seven datasets.
\FloatBarrier

\clearpage

\FloatBarrier
\section{Fitted density-interaction coefficients}
\label{app:density_coefficients}

\begin{table}[H]
\centering\small
\caption{Fitted coefficients for the density-interaction law in Equation~\ref{eq:density_interaction}, for all seven datasets and both evaluation protocols. Fits use $C\leq50$M. $C$ and $U_{\mathrm{seen}}$ are expressed in millions. Displayed coefficients are rounded; predictions use full precision.}
\label{tab:density_interaction_coefficients}
\setlength{\tabcolsep}{4pt}
\resizebox{\linewidth}{!}{%
\begin{tabular}{llrrrrrr}
\toprule
Eval. & Dataset & $L$ & $A$ & $\alpha$ & $D$ & $\beta$ & $\gamma$ \\
\midrule
kNN & Maxar & $0.91882$ & $10$ & $2.2196$ & $0.00038766$ & $2.9946$ & $1.4553$ \\
kNN & SATLAS & $0.95991$ & $0.79081$ & $0.94922$ & $0.016939$ & $0.39017$ & $0.43362$ \\
kNN & Waymo & $0.81088$ & $0.61402$ & $1.4106$ & $1.2794\times10^{-7}$ & $-5$ & $3.046$ \\
kNN & nuPlan & $0.81101$ & $0.088491$ & $0.47442$ & $7.9994\times10^{-6}$ & $2.7371$ & $5$ \\
kNN & Ego4D & $1.05$ & $0.69955$ & $1.0153$ & $0.53615$ & $0.0081791$ & $0.0064926$ \\
kNN & ImageNet & $1.05$ & $1.0667$ & $0.27203$ & $5.14\times10^{-5}$ & $0.56032$ & $5$ \\
kNN & YFCC15M & $0.62732$ & $0.94109$ & $0.8091$ & $0.013912$ & $0.78227$ & $0.69365$ \\
\midrule
Logistic & Maxar & $0.94858$ & $0.89734$ & $0.99736$ & $0.00058508$ & $2.5821$ & $0.13024$ \\
Logistic & SATLAS & $0.97594$ & $0.51044$ & $0.63647$ & $0.0060809$ & $0.92423$ & $0.55676$ \\
Logistic & Waymo & $0.84863$ & $1.7915$ & $1.9649$ & $3.0989\times10^{-6}$ & $0.28166$ & $4.9213$ \\
Logistic & nuPlan & $0.83799$ & $0.17609$ & $1.0031$ & $5.4367\times10^{-6}$ & $-0.43264$ & $5$ \\
Logistic & Ego4D & $0.57622$ & $0.59506$ & $0.88602$ & $1.1849\times10^{-6}$ & $5$ & $5.5899\times10^{-69}$ \\
Logistic & ImageNet & $0.97104$ & $0.94476$ & $0.33792$ & $2.9598\times10^{-5}$ & $0.44163$ & $5$ \\
Logistic & YFCC15M & $1.05$ & $0.77667$ & $0.18423$ & $0.036857$ & $0.66747$ & $0.54767$ \\
\bottomrule
\end{tabular}}
\end{table}
\FloatBarrier

\FloatBarrier

\section{Scaling-law fitting details and comparisons}
\label{app:scaling_fit_details}
\label{app:scaling_law_comparison}

\paragraph{Observations and weighting.}
We fit each law separately for each pretraining dataset and evaluation protocol (kNN or logistic regression). Each observation corresponds to one subset and compute budget, with downstream scores averaged across tasks as described in Section~\ref{sec:downstream_evaluation}. Only random-subset and Full observations enter the fits. Every aggregated observation receives equal weight; we do not use inverse-variance weighting or rebalance observations across compute budgets. Parameters are fitted using $C\leq50$M and evaluated using $C>50$M. We use the training quantities defined in Section~\ref{sec:sampling_strategies}, with $C$ and $U_{\mathrm{seen}}$ expressed in millions and scores expressed as fractions.

\paragraph{Fitting objective.}
Let $y_i$ and $\hat y_i(\theta)$ denote the observed and predicted scores for the $n$ fitting observations. We minimize the mean squared augmented residual,
\begin{equation}
\begin{aligned}
 J(\theta)&=\frac{1}{2n}\sum_{i=1}^{n}
 \left[(\hat y_i(\theta)-y_i)^2+100\,p(\hat y_i(\theta))^2\right],\\
 p(z)&=\max(z-1.05,0)+\max(-0.05-z,0).
\end{aligned}
\end{equation}
The second residual penalizes predictions outside $[-0.05,1.05]$ on the fitting observations. This is a soft penalty, not a hard constraint on predictions; extrapolated scores are not clipped. MAE is used for reporting, rather than as the optimization objective.

\paragraph{Parameter bounds.}
For all four main laws, $L\in[0,1.05]$ and $A\in[0,10]$. The compute-only law uses $\alpha\in[10^{-4},5]$. The data-constrained law uses $\beta\in[10^{-4},5]$ and $\tau\in[10^{-4},1000]$. The prescriptive law uses $\alpha,\delta\in[10^{-4},5]$, $B\in[0,10]$, and $\eta\in[0,5]$. The density-interaction law uses $\alpha\in[10^{-4},5]$, $D\in[0,1]$, $\beta\in[-5,5]$, and $\gamma\in[0,5]$. Thus, its support exponent may be negative, whereas its repetition exponent is nonnegative. The density ablations retain the corresponding bounds: unique and repetition penalty coefficients lie in $[0,1]$, support exponents in $[-5,5]$, and repetition exponents in $[0,5]$. These are numerical search bounds; fitted coefficients are not interpreted as independent mechanistic quantities.

\paragraph{Initialization and optimization.}
We use SciPy's bounded nonlinear least-squares solver with the trust-region reflective method, a two-point finite-difference Jacobian, and ordinary squared residuals. Each fit uses 29 initializations: three law-specific starting points based on the fitting-score range, the midpoint of the parameter bounds, and 25 uniformly sampled points within those bounds. The pseudorandom generator is initialized with seed 17 for each fit. Initial values are clipped to lie at least $10^{-9}$ inside each bound. Each initialization permits up to 25,000 function evaluations, with function, parameter, and gradient tolerances all set to $10^{-11}$. We retain the parameter vector with the smallest fitting objective across the 29 runs; held-out scores do not select the initialization or stopping point. The saved runs use SciPy 1.17.1 and NumPy 2.4.6.

\paragraph{Error reporting.}
For each dataset and evaluation protocol, fit and held-out MAE are the unweighted means of absolute prediction errors over the observations in the respective budget ranges. The uncertainty estimates associated with observed scores do not enter this calculation. All reported predictions use full-precision fitted parameters; the coefficients displayed in tables and figures are rounded for readability.

\paragraph{Why some baselines give similar predictions.}
Let $r=R-1$ denote the number of additional passes. When $r/\tau\ll1$, the effective-data law in Equation~\ref{eq:data_constrained} satisfies
\begin{equation}
U_{\mathrm{eff}}=U_{\mathrm{seen}}\left[1+r-\frac{r^2}{2\tau}
+\mathcal{O}\!\left(\frac{r^3}{\tau^2}\right)\right]\approx C,
\end{equation}
since $U_{\mathrm{seen}}(1+r)=C$. It then approaches a compute-only power law, although its fitted coefficients can differ. In the saved kNN fits, $\tau$ reaches its upper bound of 1,000 for Maxar, SATLAS, Ego4D, and YFCC15M. This reflects the fitted parameter bound; it does not establish a universal repetition-saturation scale for these datasets. The approximation can also become less accurate at the largest repetition counts.

The prescriptive penalty $Br^\delta/U_{\mathrm{seen}}^\eta$ is negligible for Maxar and YFCC15M in the saved kNN fits. For SATLAS and Ego4D, $\delta$ is close to its lower bound and $\eta$ is close to zero, making the penalty nearly constant over the observed configurations with $r>0$, while it remains zero at $r=0$. These parameter settings help explain why the baseline curves can look similar without being identical. They do not imply that the laws remain equivalent when extrapolated to other budgets or subset sizes.

Tables~\ref{tab:baseline_parameters} and~\ref{tab:baseline_parameters_logres} report the three baseline parameterizations fitted separately to kNN and logistic-regression scores. All seven main datasets are shown in the same order. Small nonzero coefficients and parameters at their bounds should not be interpreted as stable mechanistic measurements. The full comparison curves remain in Appendices~\ref{app:knn_scaling} and~\ref{app:logres_scaling}.

\FloatBarrier
\begin{table}[H]
\centering

\caption{kNN baseline parameters fitted using $C\leq50$M, rounded for display. The exponent column is $\alpha$ for compute-only and prescriptive laws and $\beta$ for the data-constrained law. Dashes indicate parameters absent from that law. $C$ and $U_{\mathrm{seen}}$ are expressed in millions; scores are fractions.}
\label{tab:baseline_parameters}
\scriptsize
\setlength{\tabcolsep}{3pt}
\resizebox{\textwidth}{!}{%
\begin{tabular}{llrrrrrrr}
\toprule
Dataset & Law & $L$ & $A$ & $\alpha$ or $\beta$ & $\tau$ & $B$ & $\delta$ & $\eta$ \\
\midrule
Maxar & Compute-only & $0.91191$ & $10$ & $2.2598$ & -- & -- & -- & -- \\
Maxar & Data-constrained & $0.91194$ & $10$ & $2.2628$ & $1000$ & -- & -- & -- \\
Maxar & Prescriptive & $0.91191$ & $10$ & $2.2598$ & -- & $7.618\!\times\!10^{-25}$ & $1\!\times\!10^{-4}$ & $5.826\!\times\!10^{-23}$ \\
\addlinespace
SATLAS & Compute-only & $0.9249$ & $0.88433$ & $1.0699$ & -- & -- & -- & -- \\
SATLAS & Data-constrained & $0.92472$ & $0.90236$ & $1.0804$ & $1000$ & -- & -- & -- \\
SATLAS & Prescriptive & $0.92989$ & $0.94726$ & $1.0976$ & -- & $0.0055379$ & $1\!\times\!10^{-4}$ & $4.54\!\times\!10^{-29}$ \\
\addlinespace
Waymo & Compute-only & $0.80985$ & $0.6791$ & $1.4692$ & -- & -- & -- & -- \\
Waymo & Data-constrained & $0.81269$ & $0.3791$ & $1.1851$ & $105.43$ & -- & -- & -- \\
Waymo & Prescriptive & $0.81086$ & $0.6128$ & $1.4105$ & -- & $1.377\!\times\!10^{-6}$ & $0.66543$ & $5$ \\
\addlinespace
nuPlan & Compute-only & $0.80089$ & $0.11817$ & $0.779$ & -- & -- & -- & -- \\
nuPlan & Data-constrained & $0.80414$ & $0.097178$ & $0.63274$ & $42.981$ & -- & -- & -- \\
nuPlan & Prescriptive & $0.80343$ & $0.10669$ & $0.68407$ & -- & $7.841\!\times\!10^{-5}$ & $1.0352$ & $3.495\!\times\!10^{-14}$ \\
\addlinespace
Ego4D & Compute-only & $0.50636$ & $0.74847$ & $1.077$ & -- & -- & -- & -- \\
Ego4D & Data-constrained & $0.50638$ & $0.74926$ & $1.0774$ & $1000$ & -- & -- & -- \\
Ego4D & Prescriptive & $0.50816$ & $0.72769$ & $1.0564$ & -- & $0.0011618$ & $1\!\times\!10^{-4}$ & $3.328\!\times\!10^{-26}$ \\
\addlinespace
ImageNet & Compute-only & $0.67539$ & $1.1558$ & $0.74651$ & -- & -- & -- & -- \\
ImageNet & Data-constrained & $1.05$ & $0.94129$ & $0.2199$ & $218.08$ & -- & -- & -- \\
ImageNet & Prescriptive & $0.80703$ & $0.98531$ & $0.48093$ & -- & $0.0010206$ & $0.74102$ & $3.293\!\times\!10^{-22}$ \\
\addlinespace
YFCC15M & Compute-only & $0.57443$ & $1.1069$ & $0.98413$ & -- & -- & -- & -- \\
YFCC15M & Data-constrained & $0.57392$ & $1.1323$ & $0.99637$ & $1000$ & -- & -- & -- \\
YFCC15M & Prescriptive & $0.57443$ & $1.1069$ & $0.98413$ & -- & $4.91\!\times\!10^{-18}$ & $1\!\times\!10^{-4}$ & $0.13335$ \\
\bottomrule
\end{tabular}}
\end{table}

\FloatBarrier
\begin{table}[H]
\centering

\caption{Logistic-regression baseline parameters fitted using $C\leq50$M, rounded for display. The exponent column is $\alpha$ for compute-only and prescriptive laws and $\beta$ for the data-constrained law. Dashes indicate parameters absent from that law. $C$ and $U_{\mathrm{seen}}$ are expressed in millions; scores are fractions.}
\label{tab:baseline_parameters_logres}
\scriptsize
\setlength{\tabcolsep}{3pt}
\resizebox{\textwidth}{!}{%
\begin{tabular}{llrrrrrrr}
\toprule
Dataset & Law & $L$ & $A$ & $\alpha$ or $\beta$ & $\tau$ & $B$ & $\delta$ & $\eta$ \\
\midrule
Maxar & Compute-only & $0.93988$ & $0.93086$ & $1.0413$ & -- & -- & -- & -- \\
Maxar & Data-constrained & $0.93855$ & $1.0056$ & $1.0829$ & $1000$ & -- & -- & -- \\
Maxar & Prescriptive & $0.93988$ & $0.93086$ & $1.0413$ & -- & $1.507\!\times\!10^{-40}$ & $1\!\times\!10^{-4}$ & $2.016\!\times\!10^{-9}$ \\
\addlinespace
SATLAS & Compute-only & $0.95418$ & $0.51062$ & $0.68413$ & -- & -- & -- & -- \\
SATLAS & Data-constrained & $0.95309$ & $0.52225$ & $0.69887$ & $1000$ & -- & -- & -- \\
SATLAS & Prescriptive & $0.95547$ & $0.52032$ & $0.69369$ & -- & $0.0019491$ & $1\!\times\!10^{-4}$ & $4.035\!\times\!10^{-29}$ \\
\addlinespace
Waymo & Compute-only & $0.8467$ & $4.0294$ & $2.3579$ & -- & -- & -- & -- \\
Waymo & Data-constrained & $0.84801$ & $1.231$ & $1.834$ & $146.65$ & -- & -- & -- \\
Waymo & Prescriptive & $0.84845$ & $2.037$ & $2.0268$ & -- & $3.444\!\times\!10^{-5}$ & $1.0308$ & $0.042859$ \\
\addlinespace
nuPlan & Compute-only & $0.83654$ & $0.21278$ & $1.1218$ & -- & -- & -- & -- \\
nuPlan & Data-constrained & $0.83743$ & $0.17638$ & $1.0211$ & $81.643$ & -- & -- & -- \\
nuPlan & Prescriptive & $0.83825$ & $0.16371$ & $0.9644$ & -- & $4.82\!\times\!10^{-7}$ & $2.5985$ & $5.188\!\times\!10^{-17}$ \\
\addlinespace
Ego4D & Compute-only & $0.57575$ & $0.59524$ & $0.88844$ & -- & -- & -- & -- \\
Ego4D & Data-constrained & $0.57579$ & $0.59572$ & $0.88865$ & $1000$ & -- & -- & -- \\
Ego4D & Prescriptive & $0.5758$ & $0.59524$ & $0.88844$ & -- & $0.024999$ & $1\!\times\!10^{-4}$ & $5$ \\
\addlinespace
ImageNet & Compute-only & $0.73982$ & $1.0566$ & $0.69924$ & -- & -- & -- & -- \\
ImageNet & Data-constrained & $1.05$ & $0.88261$ & $0.23877$ & $258.53$ & -- & -- & -- \\
ImageNet & Prescriptive & $0.85473$ & $0.92128$ & $0.46169$ & -- & $3.54\!\times\!10^{-4}$ & $0.87605$ & $8.699\!\times\!10^{-23}$ \\
\addlinespace
YFCC15M & Compute-only & $0.98082$ & $0.68704$ & $0.14545$ & -- & -- & -- & -- \\
YFCC15M & Data-constrained & $0.92337$ & $0.63856$ & $0.16767$ & $1000$ & -- & -- & -- \\
YFCC15M & Prescriptive & $0.98082$ & $0.68704$ & $0.14545$ & -- & $2.385\!\times\!10^{-57}$ & $0.076982$ & $4.531\!\times\!10^{-37}$ \\
\bottomrule
\end{tabular}}
\end{table}

\clearpage

\section{Model size and self-distillation method}
\label{app:model_method_density}
\begin{figure}[H]
    \centering
    \includegraphics[width=\textwidth,height=0.72\textheight,keepaspectratio]{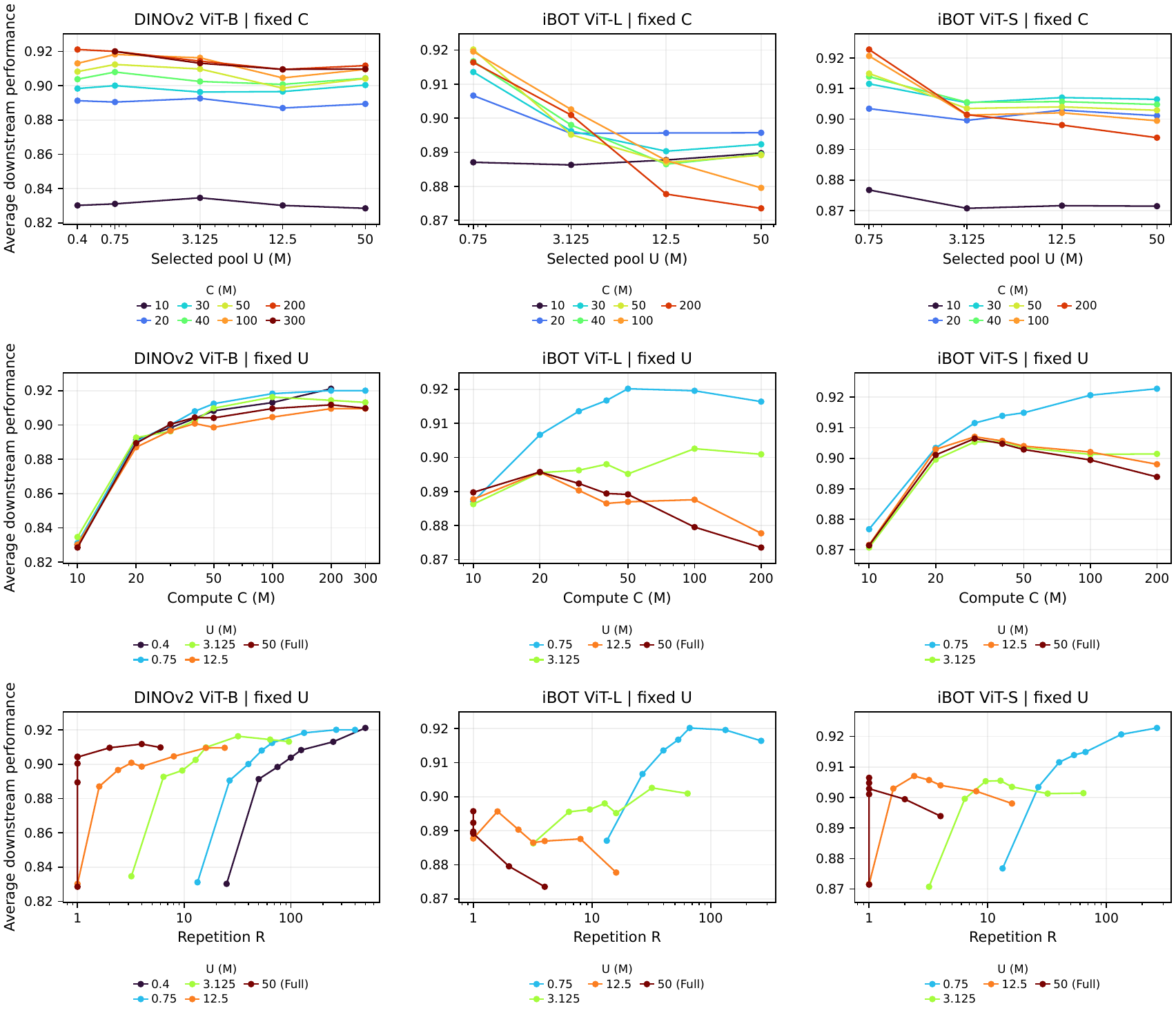}
    \caption{Maxar density curves under kNN evaluation, following the structure of Figure~\ref{fig:empirical_observations_knn}. Columns show DINOv2 with ViT-B, iBOT with ViT-L, and iBOT with ViT-S. Rows show average downstream performance versus selected pool size $U$ at fixed $C$, compute $C$ at fixed $U$, and repetition $R=C/\min(C,U)$ at fixed $U$. Points represent Full and random-subset observations; lines connect available observations. Horizontal axes are logarithmic, and $U$ and $C$ are expressed in millions.}
    \label{fig:model_method_density_knn}
\end{figure}

\clearpage
{
\section{Reduced-augmentation density curves}
\label{app:reduced_augmentation_density}
\begin{figure}[H]
    \centering
    \includegraphics[width=\textwidth]{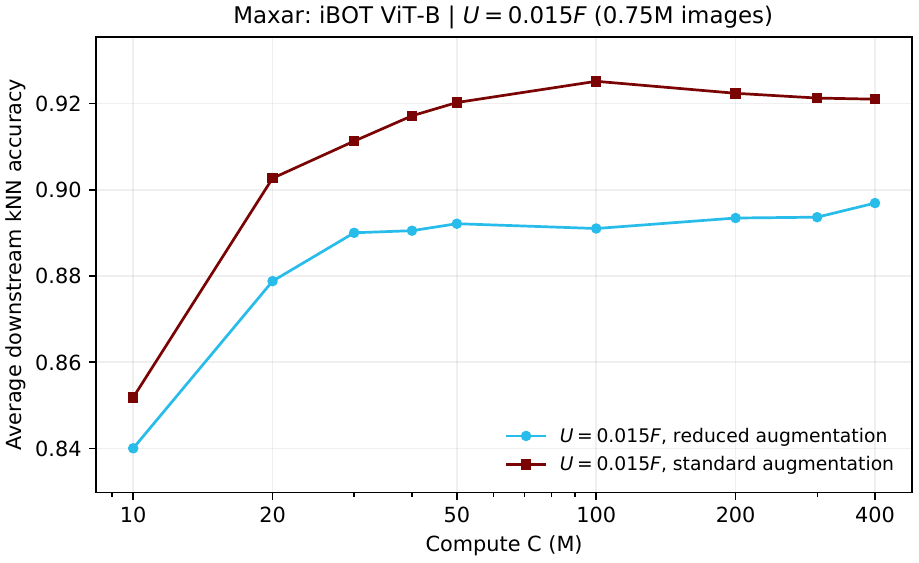}
    \caption{Average downstream kNN performance versus compute for iBOT ViT-B on Maxar, following the middle-row view of Figure~\ref{fig:empirical_observations_knn}. Both curves use a random subset with $U=0.015F$ (0.75M images), comparing standard and reduced augmentation at matched training budgets. Reduced augmentation disables color jitter, grayscale, Gaussian blur, and solarization while retaining cropping, horizontal flipping, normalization, and masking. Points show observed scores, and lines connect budgets. The horizontal axis is logarithmic, with $C$ in millions of image presentations.}
\label{fig:reduced_augmentation_density_knn}
\end{figure}

\begin{table}[htbp]
\centering\small
\caption{Average downstream logistic-regression accuracy (\%) for iBOT ViT-B trained on a random 0.75M-image Maxar subset with standard or reduced augmentation. The budget $C$ is in millions of image presentations.}
\label{tab:reduced_augmentation_logres}
\begin{tabular}{rrr}
\toprule
$C$ (M) & Standard augmentation & Reduced augmentation \\
\midrule
10 & 85.07 & 82.22 \\
20 & 90.06 & 86.68 \\
30 & 91.84 & 88.45 \\
40 & 92.93 & 88.96 \\
50 & 93.09 & 89.32 \\
100 & 93.83 & 90.08 \\
200 & 93.83 & 90.26 \\
300 & 93.78 & 90.62 \\
400 & 93.79 & 90.93 \\
\bottomrule
\end{tabular}
\end{table}
}

\newpage

\end{document}